\documentclass[sigconf, review]{acmart}

\AtBeginDocument{%
  \providecommand\BibTeX{{%
    \normalfont B\kern-0.5em{\scshape i\kern-0.25em b}\kern-0.8em\TeX}}}

\setcopyright{acmcopyright}
\copyrightyear{2021}
\acmYear{2021}
\acmDOI{10.1145/1122445.1122456}

\acmConference[CIKM '21]{CIKM '21: ACM International Conference on Information and Knowledge Management}{November 1-5, 2021}{Gold Coast, Queensland, Australia}

\acmBooktitle{CIKM '21: ACM International Conference on Information and Knowledge Management, November 1-5, 2021, Gold Coast, Queensland, Australia}

\acmPrice{15.00}
\acmISBN{978-1-4503-XXXX-X/18/06}

\usepackage{amsmath}
\usepackage{xcolor}
\usepackage{multirow}
\usepackage{makecell}
\usepackage{adjustbox}
\usepackage{tabularx}
\usepackage{enumitem}

\usepackage{color}

\usepackage{siunitx}
\usepackage{amssymb}
\usepackage{tabulary,booktabs}
\newcommand{\name}{{\texttt{FedSiam}}}

\begin{document}

\title{FedSiam: Towards Adaptive Federated Semi-Supervised Learning}


\author{Zewei Long$^+$}\thanks{$+$ This work was done when the first two authors remotely worked at Penn State University.}
\affiliation{%
  \institution{USTC}
  \country{China}
}
\email{lza@mail.ustc.edu.cn}
\author{Liwei Che$^+$}
\affiliation{%
  \institution{UESTC}
  \country{China}
}
  \email{cheliwei@std.uestc.edu.cn}
\author{Yaqing Wang}
\affiliation{%
  \institution{Purdue University}
  \country{United States}
}
\email{wang5075@purdue.edu}
\author{Muchao Ye}
\affiliation{%
  \institution{Penn State University}
  \country{United States}
}
\email{muchao@psu.edu}
\author{Junyu Luo}
\affiliation{%
  \institution{Penn State University}
  \country{United States}
}
\email{junyu@psu.edu}
\author{Jinze Wu}
\affiliation{%
  \institution{USTC}
  \country{China}
}
\email{hxwjz@mail.ustc.edu.cn}
\author{Houping Xiao}
\affiliation{%
  \institution{Georgia State University}
  \country{United States}
}
\email{hxiao@gsu.edu}
\author{Fenglong Ma$^*$}\thanks{$*$ Corresponding author.}
\affiliation{%
  \institution{Penn State University}
  \country{United States}
}
\email{fenglong@psu.edu}
%

\settopmatter{authorsperrow=4}

\settopmatter{printacmref=false} 
\pagestyle{plain} 

\begin{abstract}
Federated learning (FL) has emerged as an effective technique to co-training machine learning models without actually sharing data and leaking privacy. However, most existing FL methods focus on the supervised setting and ignore the utilization of unlabeled data. Although there are a few existing studies trying to incorporate unlabeled data into FL, they all fail to maintain performance guarantees or generalization ability in various real-world settings. In this paper, we focus on designing a general framework {\name} to tackle different scenarios of federated semi-supervised learning, including four settings in the labels-at-client scenario and two setting in the labels-at-server scenario. {\name} is built upon a siamese network into FL with a momentum update to handle the non-IID challenges introduced by unlabeled data. We further propose a new metric to measure the divergence of local model layers within the siamese network. Based on the divergence, {\name} can automatically select layer-level parameters to be uploaded to the server in an adaptive manner.
Experimental results on three datasets under two scenarios with different data distribution settings demonstrate that the proposed {\name} framework outperforms state-of-the-art baselines.
\end{abstract}




\maketitle

\section{Introduction}

Federated Learning (FL) attracts increasing attention from both academic and industrial researchers, due to its unique characteristic of collaborating in training machine learning models without actually sharing local data and leaking privacy~\cite{McMahan2017CommunicationEfficientLO,DBLP:journals/corr/abs-1902-04885,DBLP:journals/corr/abs-1912-04977}. FL has been widely applied in different applications, such as keyboard prediction~\cite{DBLP:journals/corr/abs-1811-03604}, vocal classifier~\cite{Leroy2019FederatedLF}, financial risk prediction~\cite{Yang2019FFDAF}, and medical researches~\cite{Brisimi2018FederatedLO}.
Existing FL studies~\cite{Sahu2018OnTC,Li2020OnTC,han2020robust} mainly focus on learning a global model by aggregating local model parameters, which are trained with fully labeled data. However, labeling data is expensive, time-consuming, and may need the participation of domain experts. Thus, how to utilize unlabeled data residing on local clients to learn the global model is a new challenge for FL.

Recently, federated semi-supervised learning (FedSSL) approaches ~\cite{jin2020utilizing} are proposed to tackle this challenge by integrating unlabeled data into the federated supervised learning framework, such as FedSem~\cite{Albaseer2020ExploitingUD} and FedMatch~\cite{Jeong2020FederatedSL}. FedSem employs the pseudo-labeling technique to generate fake labels for unlabeled data based on the trained FedAvg~\cite{McMahan2017CommunicationEfficientLO} model with labeled data. The data with pseudo labels are further used to retrain FedAvg to obtain the final global model. Thus, FedSem easily overfits to imperfect annotations from the pre-trained FedAvg and further leads to unsatisfactory performance. FedMatch is the state-of-the-art model for FedSSL, which introduces a new inter-client consistency loss and decomposition of the parameters learned from labeled and unlabeled data, but this approach ignores some new settings of FedSSL.

\begin{figure}[t]
\centering
\includegraphics[width=0.48\textwidth]{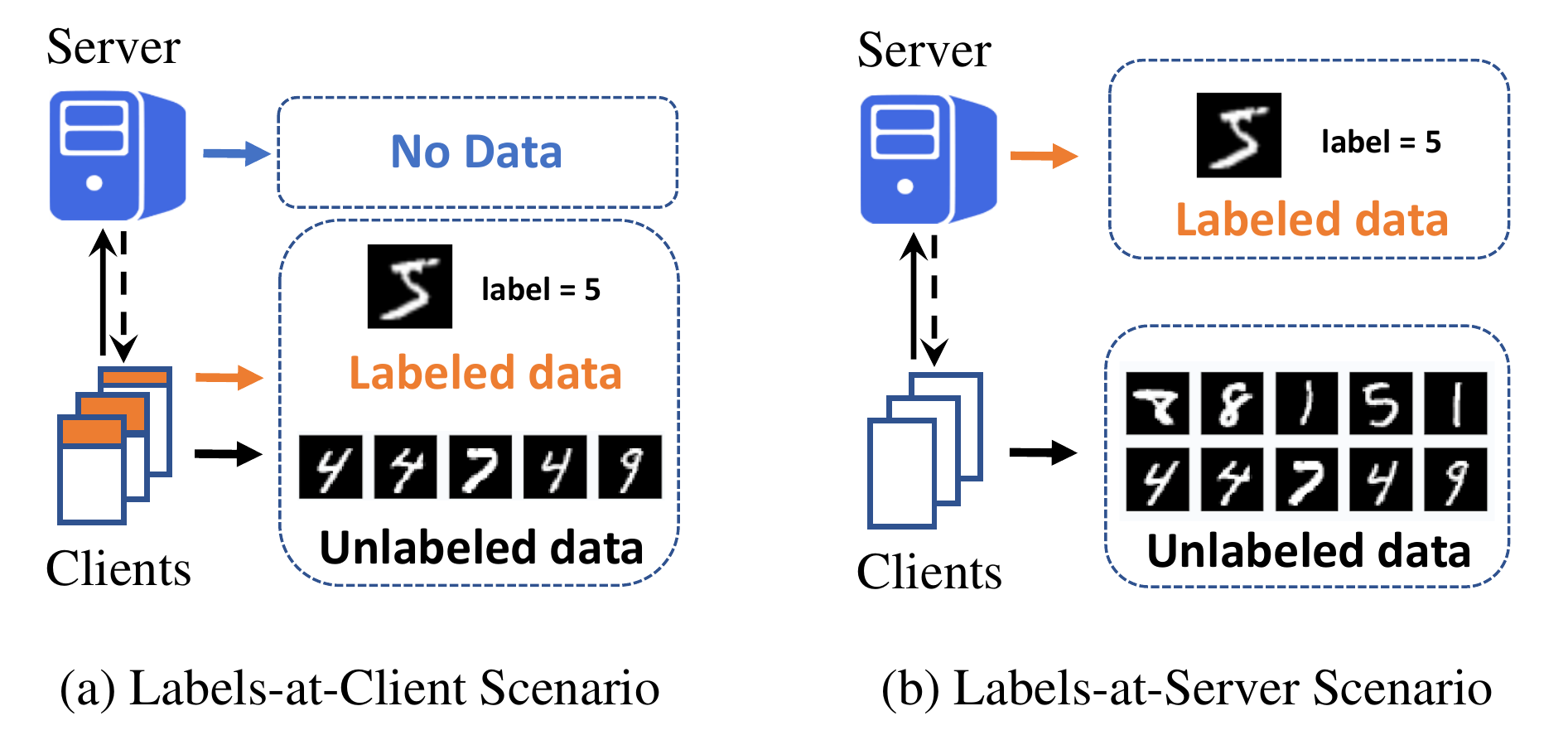} 
\vspace{-0.3in}
\caption{Illustration of two scenarios in federated semi-supervised learning. (a) Labels-at-Client scenario: both labeled and unlabeled data are available at local clients. (b) Labels-at-Server scenario: labeled data are available only at the server, while unlabeled data are available at local clients.} 
\label{fig:two_scenarios}
\vspace{-0.15in}
\end{figure}

\begin{figure*}[t]
\centering
\vspace{-0.2in}
\includegraphics[width=0.95\textwidth]{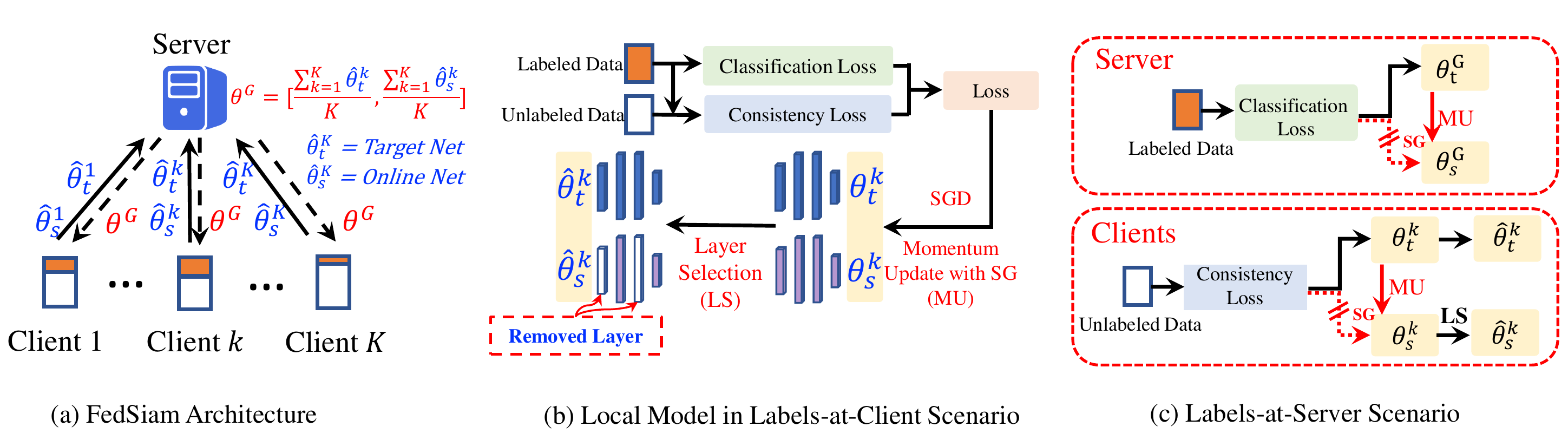} 
\caption{Overview of the proposed {\name} Framework. (a) \textit{{\name} architecture}. Each client $k$ updates the siamese network with its own data and uploads the model to the server. At the server side, the global parameters $\boldsymbol{\theta}^G$ will be updated by averaging all the local parameters and further distributed to each client. In the labels-at-server scenario, a server update for $\boldsymbol{\theta}^G$ is required before the next round. (b) \textit{Local Model for Label-at-Client Scenario}. The local model will be trained to minimize both a classification loss and a consistency loss with a momentum update (MU). The layer-level weight divergence of the siamese network is calculated. Based on a designed adaptive measure, some layers of the online net will be removed when uploading local parameters to the server. (c) \textit{Label-at-Server Scenario}. Unlike the Labels-at-Client Scenario, the Labels-at-Server Scenario only updates its local model by the consistency loss, where the layer selection (LS) and momentum update (MU) is required. Stop Gradient (SG) is a part of momentum update (MU).}

\label{framework_overivew}
\end{figure*}

\smallskip
\noindent\textbf{Motivation}.
There are two general scenarios of FedSSL, i.e., \textit{labels-at-client} and \textit{labels-at-server} as shown in Figure~\ref{fig:two_scenarios}. Labels-at-client means that there are both labeled and unlabeled data on each client, but there is no data on the server. Labels-at-server indicates that there is no labeled data on any client, and all the labeled data are stored on the server. 
A fundamental challenge of both FL and FedSSL comes from the statistical heterogeneity of data distributions, i.e, the non-IID challenges.
The current non-IID setting used by existing FedSSL studies follows that used in supervised federated learning. Assume that the number of label categories is $C$ in the whole dataset. The non-IID setting of the labels-at-client scenario assumes that there are only $C^\prime_k < C$ categories' data, including both labeled and unlabeled data, stored on the $k$-th client. However, this type of federated semi-supervised learning setting is too restricted to handle various types of data distributions in complicated real-world applications. 


\smallskip
\noindent\textbf{New Settings of Non-IID FedSSL}.
In this paper, we consider two new settings under the labels-at-client scenario by relaxing the assumption of the existing non-IID setting. First, we assume that the number of labeled data is $C^\prime_k < C$ on the $k$-th client, but the number of categories of unlabeled data is $C$. This setting is realistic since collecting unlabeled data is much easier than labeling all the categories' data. In real-world applications, the number of labeled and unlabeled data on each client may be different. Thus, in the second new setting, we assume that both labeled and unlabeled data have $C$ categories' data on each client. However, the ratios of labeled and unlabeled data varies on different clients. Therefore, it is necessary to design a new, effective, and general federated semi-supervised learning framework to handle these four non-IID settings simultaneously, i.e., two new settings for the labels-at-client scenario and two traditional non-IID setting for corresponding scenarios.

\smallskip
\noindent\textbf{{Our Solution}}.
To tackle all the aforementioned challenges, we propose a general federated semi-supervised learning framework, named {\name} as shown in Figure~\ref{framework_overivew}, which is not only effective and robust for several new FedSSL settings but also takes communication efficiency into consideration. 
{\name} firstly introduces a siamese network into federated semi-supervised learning, which has received great success in semi-supervised learning~\cite{Tarvainen2017MeanTA} and self-supervised learning~\cite{grill2020bootstrap}. The siamese network consists of two sub-networks, i.e., an online net and a target net, which enables the model to memorize the non-IID information and provide a possible solution to solve the data heterogeneity.

Except for utilizing a siamese network, we also propose to use a consistency loss that utilizes the unlabeled data and further enhances the stability of learning. This loss provides a principled way for the model to account for heterogeneity associated with partial information and the overall accuracy of federated learning in heterogeneous networks.
Besides, we introduce a new metric to measure the divergence of each local model layer. Then an adaptive layer selection approach is designed to significantly reduce the communication cost during the training process while maintaining the global performance. 

\smallskip
\noindent\textbf{Contributions}.
In summary, the main contributions of this work are as follows:
\begin{itemize}[leftmargin=*]
    \item We investigate two new settings of federated semi-supervised learning by considering different types of data heterogeneity in real world applications.
    \item We propose a general, accurate, and adaptive framework for federated semi-supervised learning, called {\name}\footnote{The source code of the proposed {\name} framework is publicly available at \url{https://anonymous.4open.science/r/fedsiam-cikm2021-B420/}}, which handles the challenge of unlabeled data by introducing a siamese network and a consistency loss. 
    \item We further introduce a new adaptive measure to automatically select local model layers during the global parameter aggregation, which is effective for reducing communication cost as well as guaranteeing model performance.
    \item We validate the proposed framework on three image datasets under both labels-at-client and labels-at-server scenarios with both IID and non-IID settings. Experimental results demonstrate the effectiveness and efficiency of the proposed {\name} framework.
\end{itemize}

\section{Preliminaries}

\subsection{Federated Learning}

Federated learning is a new collaborative learning paradigm, which aims to learn a global model without sharing local client data. Let $G$ represent the global model and $\mathcal{L} = \{l_k\}_{k=1}^K$ denote a set of local models for $K$ client\textcolor{blue}{}s. For the $k$-th client, $\mathcal{D}_L^k = \{ (\mathbf{x}^k_1,y^k_1),\cdots, (\mathbf{x}^k_n,y^k_n) \}$ represents a set of labeled data, where $\mathbf{x}^k_i$ ($i \in\{1, \cdots, n\}$) is a data instance, $y^k_i \in \{1, \cdots, C\}$ is the corresponding label, and $C$ is the number of label categories. For learning the global model $G$, federated learning usually contains two steps: \emph{local update} and \emph{parameter aggregation}.

\smallskip
\noindent
\textbf{{Local Update.}}
Federated learning algorithms, such as FedAvg, first randomly selects $B$ local models (denoted as $\mathcal{L}^B \subset \mathcal{L}$ where $B < K$) at each communication round. 
Let $\boldsymbol{\theta}^b$ represent the learned parameter set from the $b$-th local model $l_b$ using the labeled data $\mathcal{D}_L^b$, i.e., minimizing the loss function $L(\mathcal{D}_L^b)$ with stochastic gradient descent (SGD) method.

\smallskip
\noindent
\textbf{{Parameter Aggregation.}}
The average of the learned parameters $\{\boldsymbol{\theta}^1, \cdots, \boldsymbol{\theta}^B\}$ is treated as the parameter set of the global model $G$, i.e., $\boldsymbol{\theta}^G \leftarrow \frac{1}{B} \sum_{b=1}^B \boldsymbol{\theta}^b$. The global model $G$ then broadcasts $\boldsymbol{\theta}^G$ to local models, i.e, local update. This procedure is repeated until $\boldsymbol{\theta}^G$ converges.

\subsection{Federated Semi-Supervised Learning}
As shown in Figure~\ref{fig:two_scenarios}, there are two kinds federated semi-supervised learning (FedSSL) scenarios, including \textit{labels-at-client} and \textit{labels-at-server}.

\smallskip
\noindent
\textbf{{Labels-at-Client FedSSL.}}
This scenario shares the same framework with classical federated supervised learning. The main difference is that each client only annotates a small portion of their local data (i.e., 10\% of the entire data), leaving the rest of the data unlabeled. This is a common scenario for user-generated personal data, where the clients can easily annotate partial data but may not have time or motivation to label all the data (e.g. annotating faces in pictures for photo albums or social networking). 

Taking the $k$-th client as an example, we have a set of labeled data $\mathcal{D}_L^k$ as illustrated in supervised federated learning setting. Besides, we have a set of unlabeled data denoted as $\mathcal{D}_U^k = \{ \mathbf{x}^k_{n+1},\cdots, \mathbf{x}^k_{n+m}\}$, where $m$ is the number of unlabeled data. In general, $n \ll m $. In the setting of standard semi-supervised learning, we need to simultaneously minimize losses from both labeled and unlabeled data to learn the parameters $\boldsymbol{\theta}^k$ in the \emph{local update} step as follows:
\begin{equation}\label{client_ssl_loss}
\ell^k= L(\mathcal{D}_L^k) + J(\mathcal{D}_U^k),
\end{equation}
where $L(\mathcal{D}_L^k)$ is the loss for labeled data, and $J(\mathcal{D}_U^k)$ represents the loss for the unlabeled data.

Similar to supervised federated learning in the \emph{parameter aggregation} step, we can obtain $G$ using the average of $B$ selected local models after $R_g$ communication rounds as follows:
\begin{equation}\label{parameter_aggregation}
\boldsymbol{\theta}^G_{R_g}=\sum_{b=1}^{B} \frac{n^{(b)}}{\sum_{j=1}^{B} n^{(j)}} \boldsymbol{\theta}^b_{R_g}, 
\end{equation}
where $n^{(b)}$ denotes the total number of data on the $b$-th local client. 

\smallskip
\noindent
\textbf{{Labels-at-Server FedSSL.}}
Another realistic scenario assumes that labels are only available at the server side, while local clients work with unlabeled data. This is a common case of real-world applications where labeling requires expert knowledge (e.g., annotating medical images and evaluating body postures for exercises), but the data cannot be shared with the third parties due to privacy concerns.

In this scenario, $\mathcal{D}_L = \{ (\mathbf{x}_1,y_1),\cdots, (\mathbf{x}_n,y_n) \}$ represents all the labeled data, which are located on the server. In general, $n \ll |D| $, where $|D|$ is the number of training data. on each client $k$, we have a set of unlabeled data denoted as $\mathcal{D}_U^k = \{ \mathbf{x}^k_{n+1},\cdots, \mathbf{x}^k_{n+m}\}$. The overall learning procedure is similar to the labels-at-client scenario, except that we need to minimize losses from labeled data at the server side to modify the parameters $\theta^G$ after aggregation using $\ell= L(\mathcal{D}_L)$.

For the local update, we modify our loss function due to the absence of labeled data on each client, which is $\ell^k= J(\mathcal{D}_U^k)$.
Similar to standard federated semi-supervised learning in the \emph{parameter aggregation} step, we can obtain $G$ using the average of $B$ selected local models by Eq.~(\ref{parameter_aggregation}).

\section{Methodology}

To alleviate the new dilemma caused by unlabeled data, we propose {\name}, an efficient FedSSL framework as shown in Figure~\ref{framework_overivew}. A siamese network is employed as the local model to effectively handle the unlabeled data, which consists of two nets, i.e., an online net and a target net.
However, uploading the parameters of two nets to the server significantly increases the communication cost. To solve this issue, we further introduce a dynamic hyperparameter regulatory mechanism and a communication-efficient parameter selection based on the designed weight divergence. Next, we will give the details of the proposed {\name} framework.

\subsection{Siamese Network}

As described previously, {\name} uses a siamese network to utilize the unlabeled data on each client and solve the data heterogeneity. The siamese network consists of two sub-networks, i.e., an online net and a target net. For the online net $\theta_{s,q}$, we update its parameters by minimizing the loss with SGD. Then we define the target net parameters $\theta_{t,q}$ at training step $q$ as the exponential moving average (EMA) of successive $\theta_{s,q}$. Specifically, given a target decay rate $\alpha \in [0,1]$, after each training step we perform the following update, as:
\begin{equation}\label{ema}
\theta_{t,q} = \alpha \theta_{t,q-1}+ (1-\alpha) \theta_{s,q}.
\end{equation}

Siamese network has received a great success in semi-supervised learning~\cite{Tarvainen2017MeanTA} and self-supervised learning~\cite{grill2020bootstrap}. However, we now argue that the siamese network can contribute to solve the non-IID dilemma. An intuitive explanation is that the online model and target model within the siamese network have a clear division of labor. \emph{The online network keeps updating the parameter from the non-IID training data in clients; while the target network updates slowly with the momentum mechanism and reserves the long-term information from previous training.} This division of labor enables the proposed {\name} framework to extract the feature information efficiently from the non-IID data.

\subsection{Framework Design}

\subsubsection{Loss Design} 
In the proposed {\name} framework, to utilize the unlabeled data on each client, we introduce the \textbf{consistency loss} into model training. Let $\boldsymbol{\theta}^k_t$ be the parameter set of the target net. Given two perturbed inputs $\mathbf{x}^k_i + \eta$ and $\mathbf{x}^k_i + \eta^\prime$, the consistency loss disciplines the difference between the online net’s predicted probabilities $f(\mathbf{x}^k_i + \eta; \boldsymbol{\theta}^k_s)$ and the target net's predicted probabilities $f(\mathbf{x}^k_i + \eta^\prime; \boldsymbol{\theta}^k_t)$. The consistency loss is typically represented by the Mean Squared Error (MSE): 
\begin{equation}\label{mse_loss}
    J = \frac{1}{n^\prime + m^{\prime}}\sum_{j = 1}^{n^\prime + m^{\prime}} \|f(\mathbf{x}^k_j + \eta^\prime; \boldsymbol{\theta}^k_t) - f(\mathbf{x}^k_j + \eta; \boldsymbol{\theta}^k_s)\|^2
\end{equation}
or Kullback–Leibler (KL) divergence:
\begin{equation}\label{kl_loss}
    J = \frac{1}{n^\prime + m^{\prime}}\sum_{j = 1}^{n^\prime + m^{\prime}} \mathrm{KL}(f(\mathbf{x}^k_j + \eta^\prime; \boldsymbol{\theta}^k_t) \| f(\mathbf{x}^k_j + \eta; \boldsymbol{\theta}^k_s)),
\end{equation}
where $n^\prime$ denotes the total number of labeled data and their perturbations, $m^\prime$ denotes the total number of original unlabeled and perturbed unlabeled data, $f(\cdot;\cdot)$ represents a deep neural network, and $\boldsymbol{\theta}^k_s$ is the parameter set of the online net.

We apply the widely used \textbf{cross-entropy loss} as the classification loss for the labeled data, i.e.,
\begin{equation}\label{classification_loss}
    L = \frac{1}{n^\prime}\sum_{i=1}^{n^\prime} \sum_{c=1}^C p(y_i^k = c) \log f(\mathbf{x}^{k}_i; \boldsymbol{\theta}^k_s).
\end{equation}




\subsubsection{Framework Learning} The key of federated semi-supervised learning is how to utilize the unlabeled data on each client without intervening in the classification task of other clients. {\name} employs a similar loss as Eq.~(\ref{client_ssl_loss}). 
For each client $k$ in the \textit{labels-at-client} scenario, {\name} aim to update the local model as,
\begin{equation}\label{client_loss_beta}
\hat{\boldsymbol{\theta}}^k = \min_{\boldsymbol{\theta}^k \in \boldsymbol{\Theta}} L + \beta^k J,
\end{equation}
or in the \textit{labels-at-server} scenario:
\begin{equation}\label{client_loss_beta_disjoint}
\begin{array}{c}
\hat{\boldsymbol{\theta}} = \min_{\boldsymbol{\theta} \in \boldsymbol{\Theta}} L,\\
\hat{\boldsymbol{\theta}}^k = \min_{\boldsymbol{\theta}^k \in \boldsymbol{\Theta}} \beta^k J,
\end{array}
\end{equation}
where $\hat{\boldsymbol{\theta}}^k$ is the optimal parameter set of the $k$-th local model, $\hat{\boldsymbol{\theta}}$ is the optimal parameter set of the global model, $L$ is the classification loss on the labeled data, $J$ is the consistency loss on the unlabeled data, and $\beta^k$ is a dynamic hyperparameter to control the influence of the unlabeled data.

To optimize the local model, we first learn the parameters of the online net by minimizing Eq.~(\ref{client_loss_beta}) and Eq.~(\ref{client_loss_beta_disjoint}). Then we define the target net parameters $\boldsymbol{\theta}^k_t$ at training step $q$ as the
exponential moving average (EMA) of successive $\boldsymbol{\theta}^k_s$ as:
\begin{equation}\label{ema}
\boldsymbol{\theta}^k_{t,q} = \alpha^k \boldsymbol{\theta}^k_{t,q-1}+ (1-\alpha^k) \boldsymbol{\theta}^k_{s,q},
\end{equation}
where the target net weight in local $k$ at training step $q$ is composed of the moving average of the target net's weight at training step $q-1$ and the online net's weight at training step $q$. $\alpha^k$ is a dynamic hyperparameter to control the update proportion of target net. We use the ramp-up technique with an upper EMA decay during training because the online net improves quickly early in the training, and thus the target net should forget the old, inaccurate, online weights quickly.

\subsection{Layer Selection} 

Though the siamese network tightly couples target and online nets to enhance each other, in the federated learning setting, especially for the Non-IID settings, the performance of the siamese network easily degrades due to unavoidable data bias in each client. Such a challenge makes it necessary to synchronize the parameters of target and online nets across the different clients. However, full synchronization of different target and online networks easily leads to expensive communication costs. 

To solve the aforementioned challenge, we propose to remove the layers of the target net during the uploading process to the server, which are similar to those of the target net. When conducting parameter aggregation, we can borrow the removed layers of online nets from the corresponding target nets first and then use Eq.~(\ref{parameter_aggregation}) to obtain the global model $G$. In such a way, we can significantly reduce the communication cost compared with uploading all the parameters. Next, we will introduce how to calculate layer-level similarity and then describe how to remove similar layers.

\subsubsection{Layer-level Divergence}
For calculating the layer-level model divergences, we propose a new Fair-and-Square Metric (FSM), which is defined as follows: 
\begin{equation}
    \text{FSM}^k[j] = \frac{\| {\boldsymbol{\theta}}^k_t [j] - {\boldsymbol{\theta}}^k_s[j]\|}{\| {\boldsymbol{\theta}}^k_s[j]\|},
\end{equation}
where $[j]$ represents the $j$-th layer of the online or target net, and ${\left\|\cdot \right\|} $ is the Euclidean norm. \emph{Intuitively, the smaller the $\text{FSM}^k[j]$ value, the more similar the two layers of online and target nets.}  It enables us to reduce the communication cost while guaranteeing the model performance by substituting the most similar layer of the online net for the same layer of the corresponding target network.

However, the challenge here is how to determine the boundary FSM value of removing useless layers. In the FL setting, all parameters are distributed in different clients, and thus, it is impossible to manually define a threshold for each local model. Moreover, the model might have difficulty to synchronize the boundary values in asynchronous updates. To address this issue, we design a new approach to automatically estimate a global boundary value.

\subsubsection{Adaptive Boundary Estimation}
The proposed method utilizes the $\tau$ quantile of all the layer-level divergences from clients as the boundary value. Suppose that each local model $l_k$ consists of $V$ layers, which equals to the length of $\text{FSM}^k$ vector. This vector will upload to the center sever. At the $R_{g}$-th communication round or global training epoch, we can collect $V*B$ layer-level divergence values from $B$ selected clients, which is denoted as $\text{FSM}_{R_g}$. In total, the sever side stores a divergence aggregation vector $\Lambda = [\text{FSM}_1, \cdots, \text{FSM}_{R_g}]$ with length $R_{g}*V*B$ when $R_g \leq \phi_g$. To reduce the workload of the server side, we only keep $\Lambda =[\text{FSM}_{R_{g} - \phi_g + 1}, \cdots, \text{FSM}_{R_g}]$ with length $\phi_g*V*B$ if $R_g > \phi_g$. 
The $\tau$-th quantile of $\Lambda$ is considered as the estimated boundary value. 
This value will be distributed to each client. If the $j$-th value in $\text{FSM}^k$ is smaller than the $\tau$ quantile of $\Lambda$, then the corresponding layer in the online model will be not uploaded to the center sever. Though the uploading and downloading of $\Lambda$ increases the communication cost, comparing with the reduced cost from model parameters, it can be ignored. Thus, the overall communication cost can be significantly reduced with our approach.

\subsubsection{The Value of $\tau$} 
From Eq.~(\ref{client_loss_beta}), Eq.~(\ref{client_loss_beta_disjoint}) and Eq.~(\ref{ema}), we can observe that when the global training epoch $R_g$ is small, the online and target nets are relatively similar, and the consistency loss does not dominate the optimization. Thus, it is not necessary to upload the online net to the server. However, with the increase of $R_g$, the weight of the consistency loss increases, which leads to the difference between the target and online nets. At this time, we need to sacrifice communication cost and upload a part of parameters from the online net. As the training process keeps going on, the similarity of the two nets increases. Thus, the number of uploaded parameters from the online net should be decreased. To satisfy these intuitions, we design the following two choices of ramp-down curves for the hyperparameter $\tau$:

\begin{figure}[htb]
    \centering
    \begin{minipage}{0.24\textwidth}\label{fig:l_tau}
        \centering
        \includegraphics[height=1.0in]{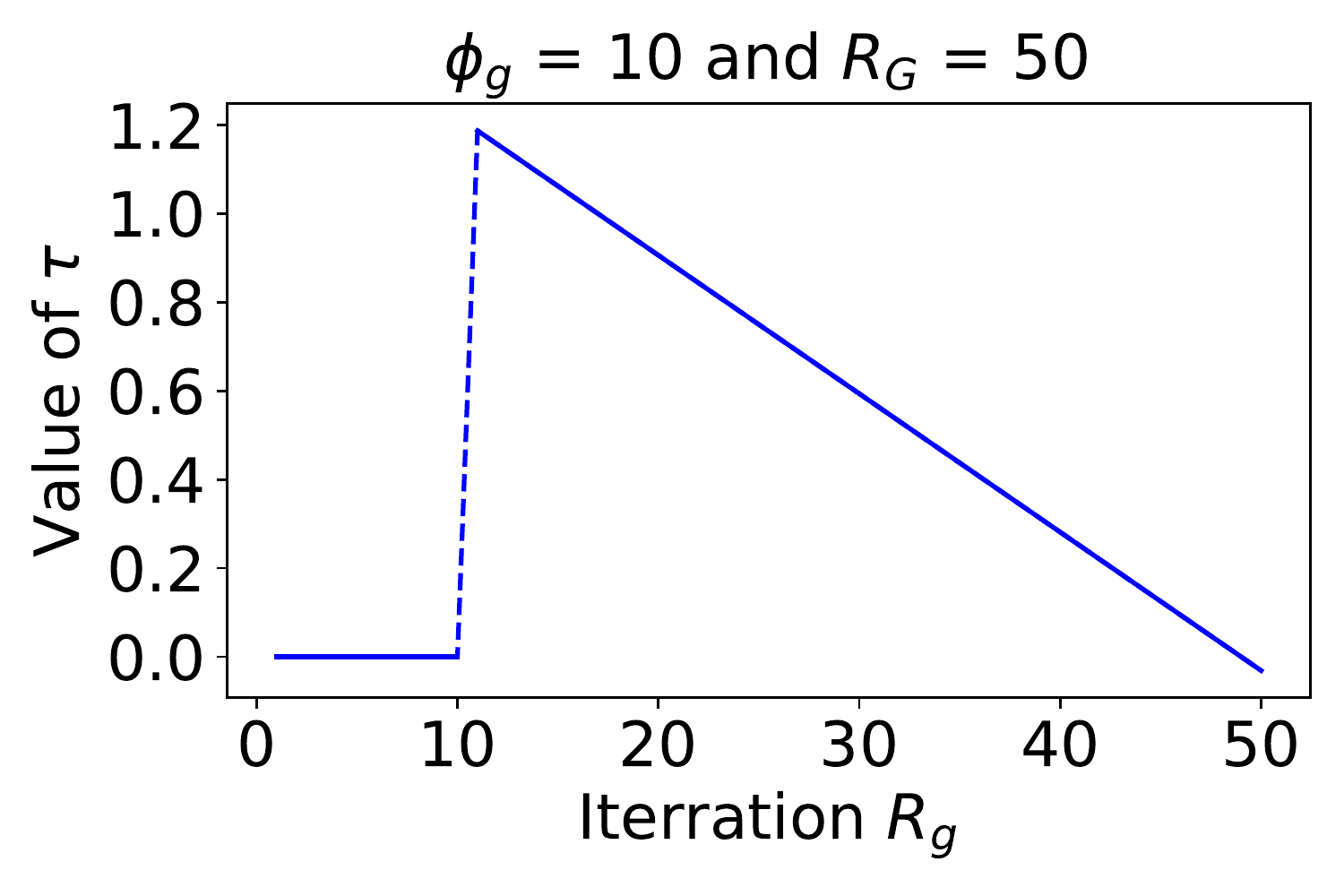}
        \vspace{-0.1in}
        \caption{Linear.}
    \end{minipage}%
    \begin{minipage}{0.24\textwidth}
        \centering
        \includegraphics[height=1.0in]{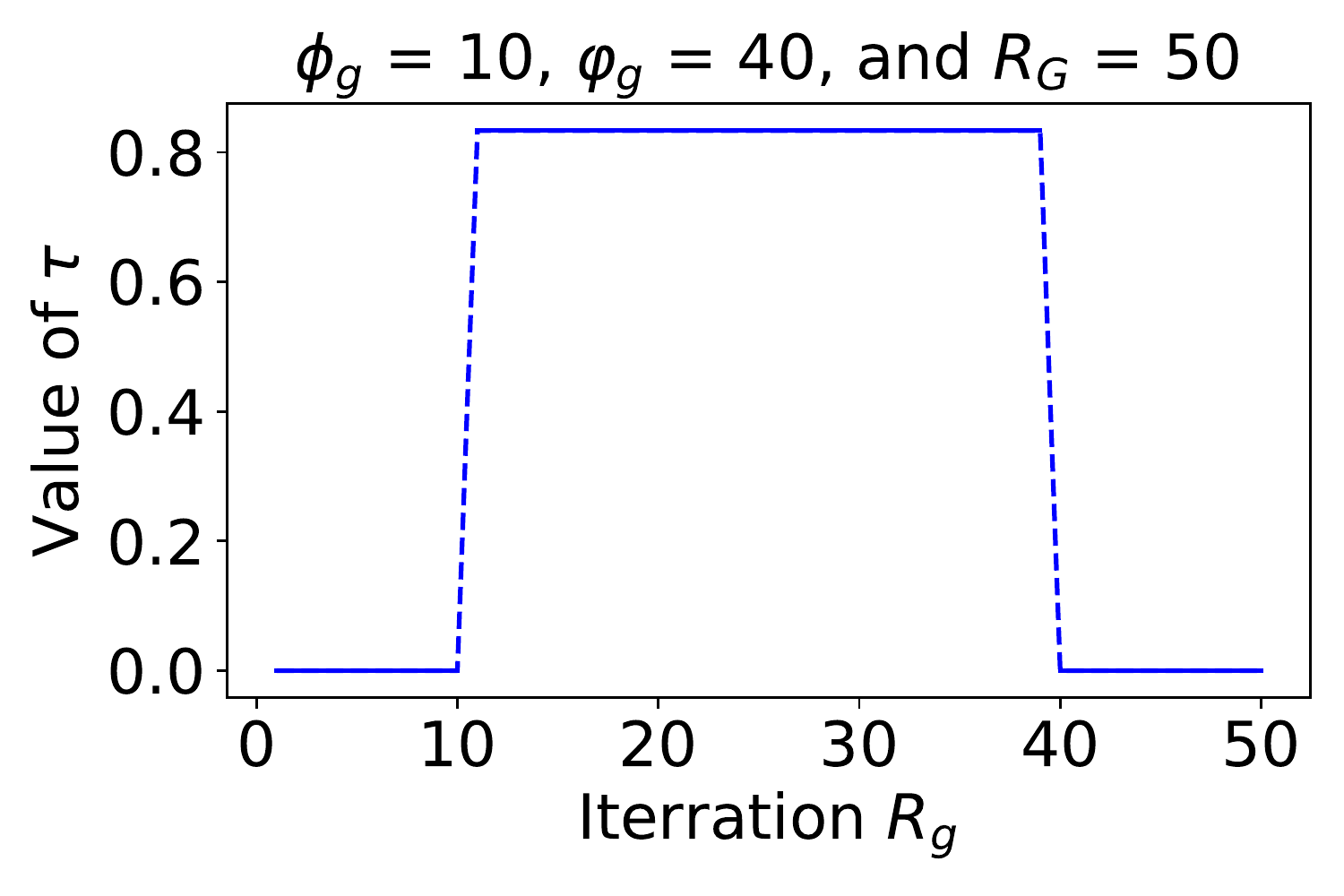}
        \vspace{-0.1in}
        \caption{Rectangle.}
    \end{minipage}
\end{figure}
\noindent\textbf{Linear quantile curve}. As shown in Figure~3, the hyperparameter $\tau$ is 0 before the training rounds $\phi_g$ and then keeps reduce to 0 linearly until the end of training.
\begin{equation}\label{l_tau}
\tau =
\begin{cases}
0,&  {R}_{g} \leq \phi_g;\\
-\frac{2* (1-\mu) * R_G}{(R_G-\phi_g)^2} (R_{g}-R_G),&  \phi_g < {R}_{g} \leq R_G.
\end{cases}
\end{equation}

\smallskip
\noindent\textbf{Rectangle quantile curve}. As shown in Figure~4, the hyperparameter $\tau$ is 0 before the training rounds $\phi_g$, then keeps as a constant until the training rounds $\varphi_g$, and equals to 0 again. 
\begin{equation}\label{r_tau}
\tau =
\begin{cases}
0,&  {R}_{g} \leq \phi_g \quad\text{or}\quad \varphi_g \leq {R}_{g} \leq R_G;\\
\frac{(1-\mu) * R_G}{\varphi_g - \phi_g},&   \phi_g < {R}_{g} < \varphi_g,
\end{cases}
\end{equation}
where $\mu \in [0, 1]$ is a predefined target of reduced communication rate compared with the siamese network, $R_{g}$ is the global epoch at that time, $\varphi_g$ is a predefined ending epoch, and $R_G$ is the total number of training epochs.

\noindent{\textbf{Remark}}. When ${R}_{g} \leq \phi_g$, we have $\tau=0$ and $\alpha^k=0$, which means that the proposed {\name} only updates the target net, which is equivalent to the $\Pi$-model. Compared with the Mean-Teacher model, the $\Pi$-model can converge quickly.
When $\phi_g < {R}_{g} \leq R_G$, indicating $\tau>0$ and $\alpha^k>0$, {\name} updates both the target and online nets, which is similar to the Mean-Teacher model.
Thus, the proposed {\name} framework is more general and fully makes use of the advantages of both state-of-the-art models to not only reduce the communication rate but also increase the model convergence rate.

\section{Experiments}

\subsection{Experimental Setup}
\noindent\textbf{Datasets}. 
In our experiments, we use three public available datasets, including MNIST\footnote{\url{http://yann.lecun.com/exdb/mnist/}}, CIFAR-10\footnote{\url{https://www.cs.toronto.edu/~kriz/cifar.html}}, and SVHN\footnote{\url{http://ufldl.stanford.edu/housenumbers/}}.
The MNIST dataset is divided into a training set of 60,000 images and a test set of 10,000 images. There are 50,000 training samples and 10,000 testing samples in the CIFAR-10 dataset. In the SVHN dataset, 73,257 digits are used for training and 26,032 digits for testing. These three datasets are all used for the image classification task with 10 categories. 

\smallskip
\noindent\textbf{Labels-at-Client Scenario}. 
Each dataset will be randomly shuffled and distributed to $K = 100$ clients, and each client will have $\frac{|D|}{K}$ instances, where $|D|$ is the number of training data. Let $\gamma$ represent the ratio of the labeled data on each client. Namely, there are $\frac{|D|}{K} * \gamma$ labeled data and $\frac{|D|}{K} * (1-\gamma)$ unlabelled data stored on each client. 
We set $\gamma = 0.1,0.15,$ and $0.2$ in the experiments. 

For the \textbf{IID} setting, both labeled and unlabeled data all have $C$ categories.
Since there are three settings under this scenario, we denote the traditional setting as \textbf{Non-IID-I} setting,  where both the labeled and unlabeled data contain 2 random categories (i.e., $C^\prime = 2$).
We also propose two new settings, which are denoted as {Non-IID-II} and {Non-IID-III}, respectively.
For the three Non-IID settings, we will use different approaches to partition the data. 
In the \textbf{Non-IID-II} setting, the labeled data have 2 categories, but the unlabeled data contain all the 10 categories. 
In the \textbf{Non-IID-III} setting, both labeled and unlabeled data all have the 10 categories. However, the different client will have different ratios of labeled data. In our experiments, 10 clients own 55\% labeled data, and 90 clients only own 5\% labeled data.

\smallskip
\noindent\textbf{Labels-at-Server Scenario}.
In this scenario, each dataset will be randomly shuffled into two parts, i.e., labeled and unlabeled data. 
Let $\gamma$ represent the ratio of the labeled data on the entire dataset. There are $|D| * \gamma$ labeled data at the server side, and $|D| * (1-\gamma)$ unlabelled data are distributed to clients, where $|D|$ is the number of training data. 
We then divided the unlabelled data into $K = 100$ clients, and each client has $\frac{|D| * (1-\gamma)}{K}$ unlabeled instances.
We set $\gamma = 0.01$ and $0.10$ in the experiments. 

For the \textbf{IID} setting, both labeled and unlabeled data all have $C$ categories. In the \textbf{Non-IID} setting, the labeled data on the server have all the 10 categories, but each client only contains 2 random categories (i.e., $C^\prime = 2$) of unlabeled data. 

\smallskip
\noindent\textbf{Baselines}.
To fairly validate the proposed {\name} framework, we use the following state-of-the-art baselines. 
\begin{itemize}[leftmargin=*]
    \item \emph{Federated semi-supervised learning models}: FedSem~\cite{Albaseer2020ExploitingUD}, FedFixMatch~\cite{Jeong2020FederatedSL}, and FedMatch~\cite{Jeong2020FederatedSL}.
    \item \emph{Federated supervised learning models}: FedAvg~\cite{McMahan2017CommunicationEfficientLO} and FedAvg$+$, which employs data augmentation techniques over FedAvg. 
\end{itemize}
These baselines will compare with the proposed {\name} framework, which has three different implements with different choices of the values of $\tau$ and $\alpha^k$, including {\name}-$\Pi$ ($\tau=0$ and $\alpha^k=0$), {\name}-MT ($\tau=1$ and $\alpha^k>0$), and {\name}-D. {\name}-D does not fix the values of $\tau$ and $\alpha^k$, which are \emph{dynamically} adjusted using Eq.~(\ref{l_tau}) or Eq.~(\ref{r_tau}). The rank of communication cost is {\name}-$\Pi < $ {\name}-D $<$ {\name}-MT.

\smallskip
\noindent\textbf{Implementation}.
For the three image datasets (including MNIST, CIFAR-10, and SVHN), we simulate the federated learning setup (1 server and $K$ devices) on a commodity machine with 1 Intel$^\circledR$ E5-2650 v4 CPU and 1 NVIDIA$^\circledR$ 2080Ti GPU. 
We use the same local model for all the baselines and {\name} on each dataset. For the MNIST dataset, we adopt a CNN~\cite{lecun1998gradient} with two 5x5 convolution layers and two linear layers (21840 total parameters). For the CIFAR-10 $\&$ SVHN datasets, we apply a CNN with six convolution layers and three linear layers (5,852,170 total parameters). The details of each model architecture are shown in Table~\ref{tab-local-arch}.

\begin{table}[!htb]
\centering
\caption{Local Model Architectures.}
\label{tab-local-arch}
\vspace{-0.1in}
\resizebox{0.47\textwidth}{!}{
\begin{tabulary}{\linewidth}{c|c|l}
\toprule
Dataset & ID & Operation \\\midrule
\multirow{4}{*}{MNIST}
& 1 & \makecell[l]{Convolution ($10 \times 5 \times 5$) + Max Pooling ($2\times2$)}\\ \cline{2-3}
& 2 & \makecell[l]{Convolution ($20 \times 5 \times 5$) + Max Pooling ($2\times2$)} \\ \cline{2-3}
& 3 & \makecell[l]{Fully Connected ($320 \times 50$) + ReLU} \\ \cline{2-3}
& 4 & \makecell[l]{Fully Connected ($50 \times 10$) + Softmax} \\ \midrule
\multirow{9}{*}{\makecell{\\CIFAR \\\& \\ SVHN}}
& 1 & \makecell[l]{Convolution ($32 \times 3 \times 3$) + BatchNorm + ReLU}\\\cline{2-3}
& 2 & \makecell[l]{Convolution ($64 \times 3 \times 3$) + ReLU + Max Pooling ($2\times2$)}\\\cline{2-3}
& 3 & \makecell[l]{Convolution ($128 \times 3 \times 3$) + BatchNorm + ReLU}\\\cline{2-3}
& 4 & \makecell[l]{Convolution ($128 \times 3 \times 3$) \\+ ReLU + Max Pooling ($2\times2$) + dropout(0.05)}\\\cline{2-3}
& 5 & \makecell[l]{Convolution ($256 \times 3 \times 3$) + BatchNorm + ReLU}\\\cline{2-3}
& 6 & \makecell[l]{Convolution ($256 \times 3 \times 3$) + ReLU +Max Pooling ($2\times2$)}\\\cline{2-3}
& 7 & \makecell[l]{Fully Connected ($4096 \times 1024$) + ReLU + Dropout (0.1)} \\\cline{2-3}
& 8 & \makecell[l]{Fully Connected ($1024 \times 512$) + ReLU + Dropout (0.1)} \\\cline{2-3}
& 9 & \makecell[l]{Fully Connected ($512 \times 10$) + Softmax} \\
\bottomrule
\end{tabulary}
}
\end{table}

For all the baselines and {\name}, we adopt the weak data argumentation technique for the three datasets, and the main process contains random reflect, flip, contrast adjustment, grayscale, and crop. 
In our experiments, there are several parameters that are shared by baselines and the proposed {\name} framework as shown in Table~\ref{tab-fl-para}.
The values of those parameters are refereed to existing works~\cite{Albaseer2020ExploitingUD, Jeong2020FederatedSL, McMahan2017CommunicationEfficientLO}. 
We use $BS_{tr}=10$ and $BS_{te}=128$ for the image classification tasks. In all experiments, we set $seed = 1234$ during dataset splitting and model training. 
\begin{table}[h]
\centering
\caption{Shared parameters for all baselines and {\name}.}
\label{tab-fl-para}
\vspace{-0.1in}
\begin{tabulary}{\linewidth}{lll}
\toprule
 \textbf{Symbol}&\textbf{Value}&\textbf{Definition}\\
\midrule
$R_G$ & 50/150/200 & round of global training\\
$K$ & 100 & total number of clients\\
$B$ & 10 & number of active clients\\
$R_L$ & 5/1 & number of local epochs \\
$BS_{tr}$ & 10/100   & local training batch size \\ 
$BS_{te}$ & 128/64   & local testing batch size \\ 
$\gamma$ & 0.01/0.1/0.15/0.2 & fraction of labeled data\\
\midrule
$lr$ & 0.01   & learning rate \\ 
$M$ & 0.9   & momentum \\ 
$wd$ & 1e-4 & weight-decay \\
$Seed$ & 1234 & random seed\\
\bottomrule
\end{tabulary}
\end{table}

Table~\ref{tab-baseline-para} shows some key parameters of baselines, including FedSem~\cite{Albaseer2020ExploitingUD}, FedFixMatch~\cite{Jeong2020FederatedSL}, and FedMatch~\cite{Jeong2020FederatedSL}. Most of those settings are the same as the primitive parameter mentioned in those works. The strong data argumentation technique is only used in FedMatch and FedFixmatch baselines.

\begin{table}[h]
\centering
\caption{Parameters for other baselines.}
\label{tab-baseline-para}
\vspace{-0.1in}
\begin{tabulary}{\linewidth}{CCl}
\toprule
 \textbf{Symbol}&\textbf{Value}&\textbf{Definition}\\
\midrule

$R_G^1$ & 30 & rounds of Phase 1 of FedSem\\
$R_G^2$ & 20 & rounds of Phase 2 of FedSem\\
$T$ & 0.95 & pseudo label threshold in Fed(Fix)Match\\
$H$ & 2 & number of helper model in FedMatch \\
$n$ & 10 & count of strong data argumentation type\\
$m$ & 10 & usage count of each argumentation type\\
\bottomrule
\end{tabulary}
\end{table}

Table~\ref{tab-fedsemi-para} shows the hyperparameters of the proposed {\name} framework. Note that the tipping point 1 of quantile curve $\phi_g = 3$ in linear quantile curve (Eq. (11)) and $\phi_g = 10$ in rectangle quantile curve (Eq. (12)). $\varphi_g$ is only used in rectangle quantile curve (Eq. (12)).
\begin{table}[h]
\centering
\caption{Hyperparameters for {\name} framework.}
\label{tab-fedsemi-para}
\vspace{-0.1in}
\begin{tabulary}{\linewidth}{lll}
\toprule
 \textbf{Symbol}&\textbf{Value}&\textbf{Definition}\\
\midrule
$\alpha^k_{max}$ & 0.999 & maximum value of moving average $\alpha^k$ \\
$\phi_l$  & 10 & coefficient threshold of consistent loss  \\ 
$\phi_g$ & 3/10 & tipping point 1 of quantile curve\\
$\varphi_g$ & 40 & tipping point 2 of rectangle quantile curve\\
$\tau_g$  & 0.5   &  communication rate of $\omega_s$\\ 
\bottomrule
\end{tabulary}
\end{table}

\begin{table*}[t]
\centering
\caption{Average accuracy of three runs on the three datasets under different settings for the labels-at-client scenario. Note that the Non-IID-III setting is different from other settings, where the ratios of labeled and unlabeled data are different on different clients.
}
\label{tab:lc}
\vspace{-0.1in}
\begin{small}
\begin{tabulary}{\linewidth}{c|l|ccc|ccc|ccc||ccc}
\toprule
 \multicolumn{2}{c|}{\textbf{Setting}} &\multicolumn{3}{c|}{\textbf{IID}}&\multicolumn{3}{c|}{\textbf{Non-IID-I}}&\multicolumn{3}{c||}{\textbf{Non-IID-II}}&\multicolumn{3}{c}{\textbf{Non-IID-III}}\\\hline
\textbf{ Ratio} &\textbf{Model} &{MNIST}&{CIFAR}&{SVHN}&{MNIST}&{CIFAR}&{SVHN}&{MNIST}&{CIFAR}&{SVHN}&{MNIST}&{CIFAR}&{SVHN}\\
\midrule
\multirow{8}{*}{$\gamma=0.10$}
&FedAvg      & 95.13\%   & 49.14\% & 79.41\% &  88.66\%   & 38.09\% & 45.84\%&  88.66\%   & 38.09\% & 45.84\%& 94.94\%   & 54.59\% & 80.12\%\\
&FedAvg$+$   & 96.20\%   & 57.37\% & 86.12\% &  92.95\%   & 40.78\% & 75.27\%&  92.95\%   & 40.78\% & 75.27\%& 96.08\%   & 57.73\% & 82.96\%\\
&FedSem      & 96.49\%   & 50.61\% & 85.70\% &  90.79\%   & 30.64\% & 52.50\%& 93.85\%   & 32.23\% & 62.28\%& 96.05\%   & 53.94\% & 82.12\%\\
&FedFixMatch & 93.19\%   & 56.47\% & 87.04\% &  86.68\%   & 41.98\% & 74.78\%& 86.80\%   & 43.49\% & 77.25\%& 92.92\%   & 55.32\% & 84.04\%\\
&FedMatch    & 94.11\%   & 55.17\% & 86.92\% &  86.61\%   & 44.88\% & 78.18\%& 88.66\%   & 45.81\% & 77.35\%& 92.91\%   & 49.64\% & 84.56\%\\\cline{2-14}
&{\name}-$\Pi$  & 97.04\%   & 60.87\% & 87.33\% & 95.35\%   & 45.78\% & 81.30\% & 94.28\%   & 48.01\% & 81.60\%& 96.79\%   & 55.05\% & 84.93\%\\
&{\name}-MT  & 97.16\% & \textbf{64.44\%} & 87.66\% &  94.95\% & 45.70\% & \textbf{81.87\%}& 94.36\% & \textbf{58.77\%} & 84.60\%& 97.00\% & \textbf{59.42\%} & 86.34\%\\
&{\name}-D   & \textbf{97.22\%} & 64.12\% & \textbf{88.76\%} & \textbf{95.61\%} & \textbf{48.11\%} & 81.61\% & \textbf{95.20\%} & 57.71\% & \textbf{84.95\%}& \textbf{97.01\%} & 59.24\% & \textbf{87.80\%}\\
\bottomrule
\multirow{8}{*}{$\gamma=0.15$}
&FedAvg      & 95.84\%   & 57.42\% & 83.60\%& 92.37\%   & 43.13\% & 58.98\% & 92.37\%   & 43.13\% & 58.98\% & -- & -- & --\\
&FedAvg$+$   & 96.77\%   & 62.37\% & 88.21\%& 92.73\%   & 42.52\% & 75.52\%& 92.73\%   & 42.52\% & 75.52\%& -- & -- & --\\
&FedSem      & 96.47\%   & 58.70\% & 87.61\%& 89.88\%   & 31.91\% & 59.87\%& 93.62\%   & 38.32\% & 57.13\% & -- & -- & --\\
&FedFixMatch & 94.29\%   & 61.66\% & 88.61\%& 89.84\%   & 45.01\% & 74.70\%& 89.54\%   & 49.81\% & 79.11\%  & -- & -- & --\\
&FedMatch    & 93.52\%   & 62.61\% & 88.88\%& 85.98\%   & 44.49\% & 78.55\%& 88.49\%   & 48.16\% & 76.51\%& -- & -- & --\\\cline{2-14}
&{\name}-$\Pi$  & 97.09\%   & 65.70\% & 89.55\%& 94.84\%   & 46.43\% & 83.01\% & 94.20\%   & 49.25\% & 82.32\%& -- & -- & --\\
&{\name}-MT  & 97.22\% & 65.70\% & \textbf{91.63\%}& 95.44\% & \textbf{47.21\%}  & \textbf{83.95\%} & \textbf{95.66\%} & \textbf{61.00\%} & 85.70\%& -- & -- & --\\
&{\name}-D   & \textbf{97.59\%} & \textbf{68.10\%} & 89.37\%& \textbf{95.89\%} & 46.40\%& 83.17\% & \textbf{95.66\%} & 57.40\% & \textbf{86.39\%}& -- & -- & --\\
\bottomrule
\multirow{8}{*}{$\gamma=0.20$}
&FedAvg      & 96.30\%   & 65.53\% & 87.57\% & 92.82\%   & 45.37\% & 68.98\%& 92.82\%   & 45.37\% & 68.98\%& -- & -- & --\\
&FedAvg$+$   & 97.25\%   & 66.51\% & 89.49\% & 92.88\%   & 48.22\% & 80.34\%& 92.88\%   & 48.22\% & 80.34\%& -- & -- & --\\
&FedSem      & 97.11\%   & 67.58\% & 89.24\% & 92.26\%   & 32.24\% & 65.75\%& 92.37\%   & 41.16\% & 69.81\% & -- & -- & --\\
&FedFixMatch & 94.68\%   & 66.54\% & 90.60\% & 90.25\%   & 45.31\% & 78.97\%& 87.44\%   & 49.62\% & 78.92\%& -- & -- & --\\
&FedMatch    & 94.92\%   & 64.73\% & 89.88\% & 86.14\%   & 46.36\% & 80.62\%& 87.70\%   & 49.41\% & 79.54\%& -- & -- & --\\\cline{2-14}
&{\name}-$\Pi$  & 97.58\%   & 68.26\% & 91.02\% & 95.03\%   & 45.84\% & 84.92\% & 95.71\%   & 45.52\% & 83.29\%& -- & -- & --\\
&{\name}-MT  & \textbf{97.60\%} & \textbf{71.79\%} & \textbf{91.36\%} & 95.60\% & \textbf{52.57\%}  & 83.72\% & \textbf{95.89\%} & \textbf{62.60\%} & 86.47\% & -- & -- & --\\
&{\name}-D   & 97.53\% & 69.87\% & 90.29\% & \textbf{96.47\%} & 47.04\%& \textbf{84.96\%} & 95.68\% & 59.45\% & \textbf{86.61\%}& -- & -- & --\\
\bottomrule
\end{tabulary}
\end{small}
\end{table*}

\subsection{Labels-at-Client Evaluation}
In this section, we evaluate the performance of the proposed {\name} framework using four settings, including one IID and three Non-IID settings. The performance of different approaches on three datasets can be found in Table~\ref{tab:lc}. We can observe that the proposed {\name} achieves the best performance compared with baselines.

\subsubsection{Performance Evaluation for the IID Setting}
FedAvg and FedAvg$+$ are two federated supervised learning approaches. From the accuracy values on the four datasets, we can observe that FedAvg$+$ performs better than FedAvg, which indicates that the data augmentation technique is powerful when the number of labeled data is limited. 
FedSem achieves the best performance among all the baselines on the MNIST dataset. 
Since the MNIST dataset has clear patterns for the digit images, the pseudo labels are also of good quality. Such a way can be treated as increasing the training data size. Thus, the accuracy of FedSem is greater than that of FedAvg$+$. However, the other two image datasets are much complicated compared with the MNIST dataset, which makes FedSem perform worse than FedAvg$+$. 

On the MNIST and CIFAR datasets, the accuracy of FedFixMatch and FedMatch is usually lower than that of FedAvg$+$, even though they all take unlabeled data into consideration but in different manners. These results illustrate that we need to design an effective way of using unlabeled data. Otherwise, the unlabeled data may degrade the performance. SVHN is a widely-used semi-supervised dataset. On this dataset, FedFixMatch and FedMatch perform better compared with FedAvg$+$. 

\begin{figure*}[!h]
\centering
\includegraphics[width=0.9\textwidth]{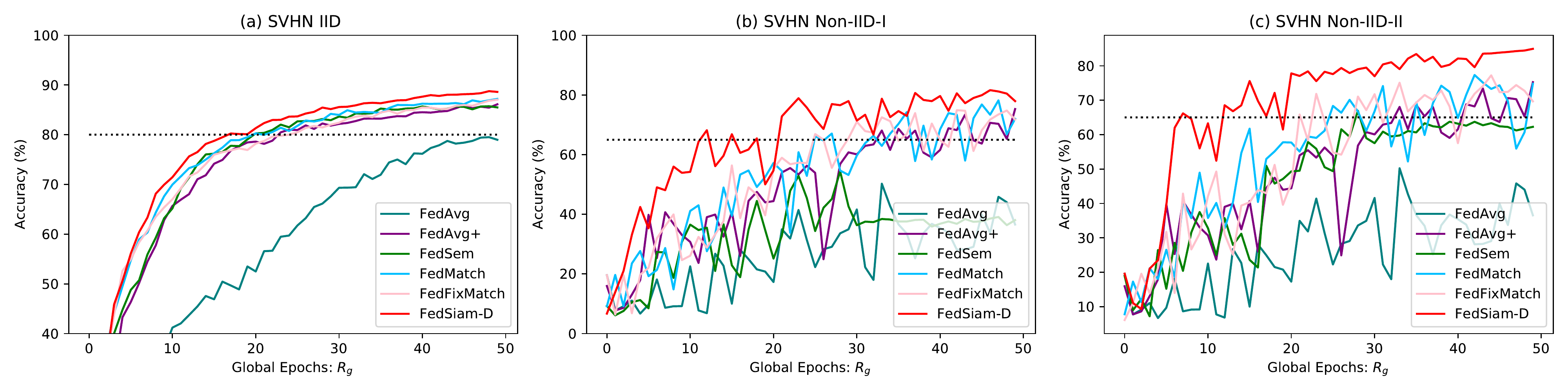} 
 \vspace{-0.15in}
\caption{Convergence curves of approaches on SVHN with $\gamma = 0.1$. Similar patterns can be observed when $\gamma = 0.15$ or $0.2$.} 
\label{SVHN-IID-convergence}
\vspace{-0.15in}
\end{figure*}

From our three implements, we can conclude that Siamese-network-based models are better than $\Pi$-based model in the IID setting. Though the model capacity of {\name}-MT is larger than that of {\name}-D, {\name}-D still achieves comparable performance with {\name}-MT on the three datasets. This demonstrates that the designed communication compression approach is reasonable and effective by dynamically adjusting the global parameter $\tau$ and the local parameter $\alpha^k$.

\subsubsection{Performance Evaluation for the Non-IID Settings}
From Table~\ref{tab:lc}, we can observe that our proposed approaches still outperform all the baselines. Compared with the results of the IID setting, we find that all the accuracy drops for both Non-IID-I and Non-IID-II settings. This observation is in accord with the fact, that is, the Non-IID setting is more challenging than the IID setting for federated learning. The accuracy of most approaches in the Non-IID-II setting is larger than that in the Non-IID-I setting. In particular, the difference of the best performance in two settings on the CIFAR-10 dataset (48.11\% and 58.77\%) is as large as 10\%. These results validate that even increasing the categories of unlabeled data may be helpful for federated semi-supervised learning.  

Under the Non-IID-III setting, we can observe that most of the accuracy values are worse than those in the IID settings, which indicates that the ratios of labeled data on different clients may hurt the performance. Compared with the results of both Non-IID-I and Non-IID-II settings, we can conclude that this imbalanced setting is easier than the previous ones, and all the approaches can increase the performance significantly.

\subsubsection{Convergence Rate} Figure~\ref{SVHN-IID-convergence} shows the test accuracy changes with respect to the number of global communication rounds or epochs $R_g$ under the IID and Non-IID settings on the SVHN dataset. We choose {\name}-D as the representative of our framework compared with all baselines. We can observe that to archive the target accuracy (i.e., the dotted line) on the SVHN dataset under all the settings, {\name}-D needs much fewer rounds but achieves greater accuracy compared with baselines. These results indicate that the fast convergence rate of {\name} also increases its efficiency.


\subsection{Labels-at-Server Evaluation}
In this section, we evaluate the performance of the proposed {\name} framework in the labels-at-server scenario, including both the IID and Non-IID settings. Since the scenario only contains unlabeled data on each client, FedAvg, FedAvg$+$, FedSem, and {\name}-D do not fit this setting. Thus, we use the remaining four approaches to validate the performance, which can be found in Table~\ref{tab:ls}. We can observe that the proposed {\name} achieves the best performance compared with baselines. 
\begin{table}[!ht]
\centering
\caption{Average accuracy of three runs on the three datasets under the labels-at-server scenario with different ratios of labeled data. }
\label{tab:ls}
\vspace{-0.1in}
\resizebox{0.47\textwidth}{!}{
\begin{tabulary}{\linewidth}{c|l|ccc|ccc}
\toprule
 \multicolumn{2}{c|}{\textbf{Setting}} &\multicolumn{3}{c|}{\textbf{IID}}&\multicolumn{3}{c}{\textbf{Non-IID}}\\\hline
\textbf{ Ratio} &\textbf{Model} &{MNIST}&{CIFAR}&{SVHN}&{MNIST}&{CIFAR}&{SVHN}\\
\midrule
\multirow{5}{*}{$\gamma=0.01$}
&FedFixMatch & 88.67\%   & 49.75\% & 75.05\% &  89.27\%   & 48.93\% & 75.92\%\\
&FedMatch    & 90.16\%   & 52.64\% & 78.52\% &  89.61\%   & 52.65\% & 77.84\%\\
\cline{2-8}
&{\name}-$\Pi$  & \textbf{91.28\%}   & 52.70\% & \textbf{80.10\%} & \textbf{91.61\%}   & 50.77\% & 80.69\%\\
&{\name}-MT  & 91.25\% & \textbf{53.90\%} & 77.27\% &  90.93\% & \textbf{55.26\%} & \textbf{81.51\%}\\
\bottomrule
\multirow{5}{*}{$\gamma=0.10$}
&FedFixMatch & 95.97\%   & 74.75\% & 92.10\% &  96.15\%   & 75.88\% & 92.08\%\\
&FedMatch    & 96.09\%   & 80.16\% & 92.09\% &  95.70\%   & 79.50\% & 91.46\%\\\cline{2-8}
&{\name}-$\Pi$  & \textbf{96.18\%}   & 80.19\% & 92.13\% & \textbf{96.32\%}   & \textbf{80.10\%} & 92.18\%\\
&{\name}-MT  & 95.82\% & \textbf{80.50\%} & \textbf{92.32\%} &  96.07\% & 79.61\% & \textbf{92.33\%}\\
\bottomrule
\end{tabulary}
}
\end{table}

\subsubsection{Performance Evaluation for the IID Setting}
On all three datasets, the accuracy of FedMatch is higher than that of FedFixMatch, because it introduces two helper networks and parameter deposition strategy in the labels-at-client scenario. These results illustrate that fixMatch-based methods can have a great performance with some new mechanisms. 

From our implements, we can conclude that siamese network-based models are better than fixmatch-based models in the IID setting of labels-at-server scenario. On all three datasets, {\name}-$\Pi$ and {\name}-MT achieve better performance compared with other baselines. When the radio of labeled data $\gamma$ = 0.10, the advantage of our proposed {\name} framework is more obvious. This demonstrates that {\name} have a great capacity for the situation that labeled data are plentiful, and data distribution is even in the labels-at-server scenario.

\subsubsection{Performance Evaluation for the Non-IID Setting}
For the Non-IID setting with $\gamma$ = 0.10, the accuracy of FedFixMatch in MNIST and SVHN is lower than that of FedMatch, even though it brings the helper networks and parameter deposition strategy into the basic fixmatch-based method in the labels-at-client scenario. These results illustrate that we need to design an effective way of using unlabeled data while considering the effect of the Non-IID setting. Otherwise, the data heterogeneity may degrade the model performance.

From Table~\ref{tab:ls}, we can observe that our {\name} outperforms other baselines in labels-at-server scenario. On all three datasets, {\name}-$\Pi$ and {\name}-MT achieve up to 4\% performance gain compared with other baselines. The advantage of our framework is more obvious when $\gamma$ = 0.01. This demonstrates that {\name} have greater performance gain in the Non-IID setting when the labeled data are extremely limited. For the SVHN dataset, {\name}-MT model perform better than {\name}-$\Pi$. We can conclude that the siamese network has more advantages for dealing with classical semi-supervised learning datasets.

\subsection{{Ablation Studies}}
In this section, we conduct experiments to analyze the importance of various aspects of {\name} under the labels-at-client scenario. We conduct experiments with $\gamma =0.1$, varying one or a few hyperparameters at a time while keeping the others fixed.

\begin{figure}[!htb]
\centering
\includegraphics[width=1\columnwidth]{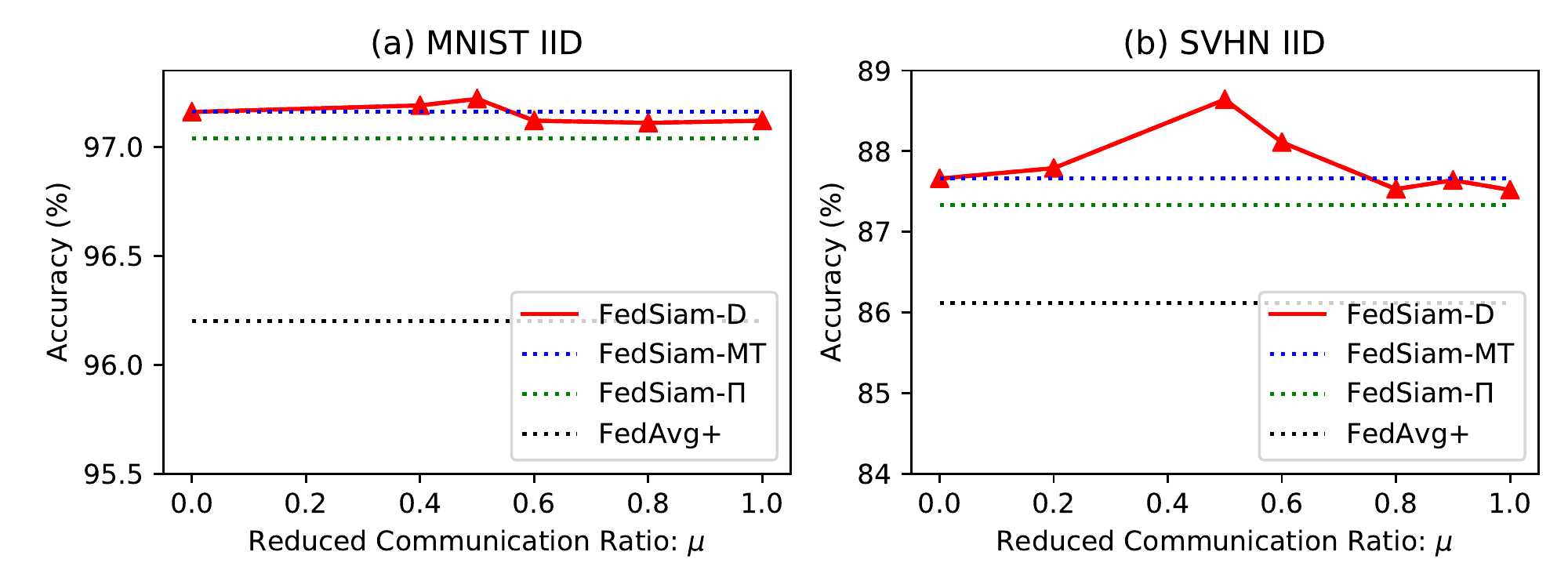}
\vspace{-0.2in}
\caption{Test accuracy v.s. $\mu$ for on the MNIST and SVHN datasets under the IID setting when $\gamma = 0.10$. }
\label{ccp}
\end{figure}

\smallskip
\noindent\textbf{Communication Efficiency}.
One benefit of the proposed {\name} framework is that it can reduce model communication cost in an adaptive manner. To explore the relationship between communication cost and accuracy, we conduct experiments by changing the hyperparameter $\mu$, which represents the target reduced communication cost compared with the full siamese network. {\name}-MT is built upon the full siamese network, which has the largest communication cost compared with FedAvg$+$, {\name}-$\Pi$, and {\name}-D. FedAvg$+$ and {\name}-$\Pi$ have the lowest communication cost. 
Figure~\ref{ccp} shows the results on the two datasets under the IID setting. The X-axis represents the value of $\mu$, and Y-axis denotes the value of accuracy.

We can observe that with the decrease of communication cost, i.e, increasing the value of $\mu$, the accuracy obtained by {\name}-D increases until achieving the peak value, which is larger than that of FedAvg$+$, {\name}-$\Pi$, and even {\name}-MT. A interesting result is that reducing the network communication ({\name}-D) even leads to performance improve compared with {\name}-MT. We can understand the layer's removal and splicing as the random noise added into the model during the training, which can contribute to the robustness and performance of our model in some cases. This shows that the proposed dynamically adaptive layer selection mechanism is reasonable and essential. After the peak value, the accuracy of {\name}-D then drops, which shows that there exists a trade-off between reducing communication cost and persevering performance. 

\smallskip
\noindent\textbf{Siamese Network \& Momentum Update}. We firstly experiment with the siamese network and momentum update, which controls the training of models. We report the performance of {\name}-$NoSiam$, i.e, without using the siamese network, as shown in the second row of Table~\ref{tab:as}. The result suggests that the siamese network plays a important role with regard to model performance. An essential hyperparameter of {\name} is the EMA decay on training $\alpha^k$. We conduct experiments to select the best value of $\alpha^k$ and validate the sensitivity of our model to these values. From the third row of Table~\ref{tab:as}, we can see that the model accuracy first increases and then degrades quickly as the $\alpha^k$ grows. According to these results, we use $\alpha^k=0.999$ in each training run, which receives the best performance among all the settings.

\smallskip
\noindent\textbf{Quantile Curve for $\tau$}. We experiment with the ramp-up curve for $\tau$, which controls the amount of unloaded parameters. We report the accuracy value of two quantile curves in the last row of Table~\ref{tab:as}, with total 50 rounds training with the IID case in the label-at-client scenario. The result demonstrates that {\name}-D achieves the best performance with the linear quantile curve on the MNIST and SVHN datasets, while the model with rectangle curve performs the best among the CIFAR-10 dataset, but the gap is small compared with the linear version. With these results, we use the linear quantile curve for the MNIST and SVHN datasets and rectangle curve for the CIFAR-10 dataset.

\begin{table}[!ht]
\centering
\caption{Mean accuracy with $\gamma=0.1$ under the IID setting over four runs per hyperparameter setting. In each experiment, we vary one hyperparameter and fix other hyperparameters listed in Table~\ref{tab-fl-para}. }
\label{tab:as}
\vspace{-0.1in}
\resizebox{0.47\textwidth}{!}{
\begin{tabulary}{\linewidth}{c|l|ccc}
\toprule
\textbf{} &\textbf{Settings} &{MNIST}&{CIFAR}&{SVHN}\\
\midrule
\multirow{2}{*}{Siamese Network}
&{\name}-NoSiam & 97.03\%   & 60.78\% & 87.24\% \\
&{\name}-D    & \textbf{97.22\%}   & \textbf{64.12\%} & \textbf{88.76\%} \\
\midrule
\multirow{4}{*}{Momentum Update $\alpha^k$}
&0.9 & 97.08\%   & 63.86\% & 88.22\% \\
&0.99    & 96.93\%   & 63.48\% & 88.25\% \\
&0.999 & \textbf{97.22\%}   & \textbf{64.12\%} & \textbf{88.76\%} \\
&0.9999    & 96.94\%   & 63.13\% & 88.60\% \\
\midrule
\multirow{2}{*}{Quantile Curve for $\tau$}
&linear & \textbf{97.22\%}   & 64.08\% & \textbf{88.76\% }\\
&rectangle   & 97.13\%   & \textbf{64.12\% }& 88.63\% \\
\bottomrule
\end{tabulary}
}
\end{table}


\smallskip
\noindent\textbf{Loss Function}. To choose the suitable loss function as the consistency loss $J$ under the FedSSL scenario, we conduct the ablation experiments on two different losses, i.e., Mean Squared Error (MSE) loss and Kullback-Leibler (KL) Divergence loss in three datasets. From Table~\ref{tab:as}, we can observe that under the IID setting, the models with MSE loss and KL Divergence loss perform better than baselines listed in Table~\ref{tab:lc}, and using MSE loss leads to the best accuracy on the three datasets. Thus, we choose to use the MSE loss for the IID setting. However, in the Non-IID-I setting, using KL Divergence loss can achieve much higher accuracy compared with the MSE loss on the CIFAR dataset, and on the MNIST and SVHN datasets, the modes with MSE loss still outperform the ones with KL Divergence loss. Based on these observations, we choose the KL Divergence loss for the CIFAR dataset and the MSE loss for both MNIST and SVHN datasets when running experiments under the Non-IID settings.

\begin{table}[!ht]
\centering
\caption{Mean accuracy with $\gamma=0.1$ under the IID and Non-IID-I setting over four runs per hyperparameter setting. In each experiment, we change the consistency losses and fix other hyperparameters listed in Table~\ref{tab-fl-para}. }
\label{tab:as}
\vspace{-0.1in}
\resizebox{0.47\textwidth}{!}{
\begin{tabulary}{\linewidth}{c|l|ccc}
\toprule
\textbf{} &\textbf{Settings} &{MNIST}&{CIFAR}&{SVHN}\\
\midrule
\multirow{2}{*}{IID}
&MSE loss & \textbf{97.22\%}   & \textbf{64.12\%} & \textbf{88.76\%} \\
&KL Divergence loss   & 96.16\%   & 63.52\% & 88.71\% \\
\midrule
\multirow{2}{*}{Non-IID-I}
&MSE loss & \textbf{95.61\%}   & 41.09\% & \textbf{81.61\%} \\
&KL Divergence loss   & 87.04\%   & \textbf{48.11\%} & 73.69\% \\
\bottomrule
\end{tabulary}
}
\end{table}

\section{Related Work}
\subsection{Federated Learning}
Federated learning aims to collaboratively build a joint model through data from different parties or clients. Most algorithms of FL focus on the supervised setting and mainly solving three challenges: statistical heterogeneity~\cite{DBLP:journals/corr/abs-1806-00582,DBLP:journals/corr/abs-1812-06127,DBLP:journals/corr/abs-1811-12629}, system constraints~\cite{Caldas2018ExpandingTR,Wang2019CMFLMC,Chen2018FederatedMW}, and trustworthiness~\cite{Bhowmick2018ProtectionAR,Geyer2017DifferentiallyPF,Bonawitz2016PracticalSA}. In this paper, we mainly aim to solve the challenge of statistical heterogeneity, i.e., the Non-IID setting.
To address this challenge, various algorithms have been proposed like sharing a some part of data ~\cite{DBLP:journals/corr/abs-1806-00582}, training personal model for each client~\cite{DBLP:journals/corr/abs-1812-06127}, or adjusting the SGD convergence of FL \cite{DBLP:journals/corr/abs-1811-12629}. 


However, introducing unlabeled data into federated learning significantly increases the difficulty of the Non-IID setting.  
Regarding FL in the semi-supervised scenario, relatively little attention has been paid to this area. A simple two-phase training with pseudo labeling is introduced to FL~\cite{Albaseer2020ExploitingUD}. A study on inter-client consistency suggested that a simple application of SSL methods might not perform well in FL, and the inter-client level consistency might improve the performance~\cite{Jeong2020FederatedSL}. However, these efforts do not propose a more general and practical algorithm and validation of its potentials on new challenges of federated semi-supervised learning.



\subsection{Semi-supervised Learning}
Semi-supervised learning mitigates the requirement for labeled data by providing a means of leveraging unlabeled data~\cite{4787647}. 
The recent works in SSL are diverse but a trend of unity. Pseudo label, which converts the unlabeled data to labeled data, utilizes unlabeled data by labeling the data with a dynamic threshold~\cite{Lee2013PseudoLabelT}. A nature and well-working idea on consistency regularization has been widely adopted in SSL~\cite{Rasmus2015SemisupervisedLW,Tarvainen2017MeanTA,Laine2017TemporalEF,Miyato2019VirtualAT,park2017adversarial}. A further discussion on how loss geometry interacts with training procedures suggests that the flat platform of SGD leads to the convergence dilemma of consistency-based SSL~\cite{Athiwaratkun2019ThereAM}. By exploring further or mixing many practical methods, UDA ~\cite{Xie2019UnsupervisedDA}, MixMatch~\cite{Berthelot2019MixMatchAH} , ReMixMatch ~\cite{Berthelot2019ReMixMatchSL}, and Fixmatch ~\cite{Sohn2020FixMatchSS} are proposed. In our work, we mainly focus on utilizing the pure consistency-based methods working with federated learning.

\section{Conclusion}
In this work, we focus on the practical and challenging setting in federated semi-supervised learning (FedSSL). To fully consider the new fundamental challenges causing by unlabeled data, we introduce two new non-IID settings in the labels-at-client scenario. Correspondingly, we propose a novel and general framework, called {\name}, which is not only effective and robust for several new FedSSL Non-IID scenarios but also takes communication efficiency into consideration. Experiments on three image datasets under the IID and Non-IID settings in both labels-at-client and labels-at-server scenarios demonstrate the effectiveness of the proposed {\name} framework compared with state-of-the-art baselines for the federated semi-supervised learning task.


\bibliographystyle{ACM-Reference-Format}
\bibliography{references.bib}


\begin{thebibliography}{36}


\ifx \showCODEN    \undefined \def \showCODEN     #1{\unskip}     \fi
\ifx \showDOI      \undefined \def \showDOI       #1{#1}\fi
\ifx \showISBNx    \undefined \def \showISBNx     #1{\unskip}     \fi
\ifx \showISBNxiii \undefined \def \showISBNxiii  #1{\unskip}     \fi
\ifx \showISSN     \undefined \def \showISSN      #1{\unskip}     \fi
\ifx \showLCCN     \undefined \def \showLCCN      #1{\unskip}     \fi
\ifx \shownote     \undefined \def \shownote      #1{#1}          \fi
\ifx \showarticletitle \undefined \def \showarticletitle #1{#1}   \fi
\ifx \showURL      \undefined \def \showURL       {\relax}        \fi
\providecommand\bibfield[2]{#2}
\providecommand\bibinfo[2]{#2}
\providecommand\natexlab[1]{#1}
\providecommand\showeprint[2][]{arXiv:#2}

\bibitem[\protect\citeauthoryear{Albaseer, Ciftler, Abdallah, and
  Al-Fuqaha}{Albaseer et~al\mbox{.}}{2020}]%
        {Albaseer2020ExploitingUD}
\bibfield{author}{\bibinfo{person}{Abdullatif Albaseer},
  \bibinfo{person}{Bekir~Sait Ciftler}, \bibinfo{person}{Mohamed Abdallah},
  {and} \bibinfo{person}{Ala Al-Fuqaha}.} \bibinfo{year}{2020}\natexlab{}.
\newblock \showarticletitle{Exploiting Unlabeled Data in Smart Cities using
  Federated Learning}.
\newblock \bibinfo{journal}{\emph{2020 International Wireless Communications
  and Mobile Computing (IWCMC)}} (\bibinfo{year}{2020}).
\newblock


\bibitem[\protect\citeauthoryear{Athiwaratkun, Finzi, Izmailov, and
  Wilson}{Athiwaratkun et~al\mbox{.}}{2019}]%
        {Athiwaratkun2019ThereAM}
\bibfield{author}{\bibinfo{person}{Ben Athiwaratkun}, \bibinfo{person}{Marc
  Finzi}, \bibinfo{person}{Pavel Izmailov}, {and}
  \bibinfo{person}{Andrew~Gordon Wilson}.} \bibinfo{year}{2019}\natexlab{}.
\newblock \showarticletitle{There are many consistent explanations of unlabeled
  data: Why you should average}.
\newblock \bibinfo{journal}{\emph{ICLR}} (\bibinfo{year}{2019}).
\newblock


\bibitem[\protect\citeauthoryear{Berthelot, Carlini, Cubuk, Kurakin, Sohn,
  Zhang, and Raffel}{Berthelot et~al\mbox{.}}{2019a}]%
        {Berthelot2019ReMixMatchSL}
\bibfield{author}{\bibinfo{person}{David Berthelot}, \bibinfo{person}{Nicholas
  Carlini}, \bibinfo{person}{Ekin~D Cubuk}, \bibinfo{person}{Alex Kurakin},
  \bibinfo{person}{Kihyuk Sohn}, \bibinfo{person}{Han Zhang}, {and}
  \bibinfo{person}{Colin Raffel}.} \bibinfo{year}{2019}\natexlab{a}.
\newblock \showarticletitle{Remixmatch: Semi-supervised learning with
  distribution matching and augmentation anchoring}. In
  \bibinfo{booktitle}{\emph{ICLR}}.
\newblock


\bibitem[\protect\citeauthoryear{Berthelot, Carlini, Goodfellow, Papernot,
  Oliver, and Raffel}{Berthelot et~al\mbox{.}}{2019b}]%
        {Berthelot2019MixMatchAH}
\bibfield{author}{\bibinfo{person}{David Berthelot}, \bibinfo{person}{Nicholas
  Carlini}, \bibinfo{person}{Ian Goodfellow}, \bibinfo{person}{Nicolas
  Papernot}, \bibinfo{person}{Avital Oliver}, {and} \bibinfo{person}{Colin~A
  Raffel}.} \bibinfo{year}{2019}\natexlab{b}.
\newblock \showarticletitle{Mixmatch: A holistic approach to semi-supervised
  learning}. In \bibinfo{booktitle}{\emph{Advances in Neural Information
  Processing Systems}}. \bibinfo{pages}{5049--5059}.
\newblock


\bibitem[\protect\citeauthoryear{Bhowmick, Duchi, Freudiger, Kapoor, and
  Rogers}{Bhowmick et~al\mbox{.}}{2018}]%
        {Bhowmick2018ProtectionAR}
\bibfield{author}{\bibinfo{person}{Abhishek Bhowmick}, \bibinfo{person}{John
  Duchi}, \bibinfo{person}{Julien Freudiger}, \bibinfo{person}{Gaurav Kapoor},
  {and} \bibinfo{person}{Ryan Rogers}.} \bibinfo{year}{2018}\natexlab{}.
\newblock \showarticletitle{Protection against reconstruction and its
  applications in private federated learning}.
\newblock \bibinfo{journal}{\emph{arXiv preprint arXiv:1812.00984}}
  (\bibinfo{year}{2018}).
\newblock


\bibitem[\protect\citeauthoryear{Bonawitz, Ivanov, Kreuter, Marcedone, McMahan,
  Patel, Ramage, Segal, and Seth}{Bonawitz et~al\mbox{.}}{2016}]%
        {Bonawitz2016PracticalSA}
\bibfield{author}{\bibinfo{person}{Keith Bonawitz}, \bibinfo{person}{Vladimir
  Ivanov}, \bibinfo{person}{Ben Kreuter}, \bibinfo{person}{Antonio Marcedone},
  \bibinfo{person}{H~Brendan McMahan}, \bibinfo{person}{Sarvar Patel},
  \bibinfo{person}{Daniel Ramage}, \bibinfo{person}{Aaron Segal}, {and}
  \bibinfo{person}{Karn Seth}.} \bibinfo{year}{2016}\natexlab{}.
\newblock \showarticletitle{Practical secure aggregation for federated learning
  on user-held data}.
\newblock \bibinfo{journal}{\emph{arXiv preprint arXiv:1611.04482}}
  (\bibinfo{year}{2016}).
\newblock


\bibitem[\protect\citeauthoryear{Brisimi, Chen, Mela, Olshevsky, Paschalidis,
  and Shi}{Brisimi et~al\mbox{.}}{2018}]%
        {Brisimi2018FederatedLO}
\bibfield{author}{\bibinfo{person}{Theodora~S Brisimi}, \bibinfo{person}{Ruidi
  Chen}, \bibinfo{person}{Theofanie Mela}, \bibinfo{person}{Alex Olshevsky},
  \bibinfo{person}{Ioannis~Ch Paschalidis}, {and} \bibinfo{person}{Wei Shi}.}
  \bibinfo{year}{2018}\natexlab{}.
\newblock \showarticletitle{Federated learning of predictive models from
  federated electronic health records}.
\newblock \bibinfo{journal}{\emph{International journal of medical
  informatics}}  \bibinfo{volume}{112} (\bibinfo{year}{2018}),
  \bibinfo{pages}{59--67}.
\newblock


\bibitem[\protect\citeauthoryear{Caldas, Kone{\v{c}}ny, McMahan, and
  Talwalkar}{Caldas et~al\mbox{.}}{2018}]%
        {Caldas2018ExpandingTR}
\bibfield{author}{\bibinfo{person}{Sebastian Caldas}, \bibinfo{person}{Jakub
  Kone{\v{c}}ny}, \bibinfo{person}{H~Brendan McMahan}, {and}
  \bibinfo{person}{Ameet Talwalkar}.} \bibinfo{year}{2018}\natexlab{}.
\newblock \showarticletitle{Expanding the reach of federated learning by
  reducing client resource requirements}.
\newblock \bibinfo{journal}{\emph{arXiv preprint arXiv:1812.07210}}
  (\bibinfo{year}{2018}).
\newblock


\bibitem[\protect\citeauthoryear{Chapelle, Scholkopf, and Zien}{Chapelle
  et~al\mbox{.}}{2009}]%
        {4787647}
\bibfield{author}{\bibinfo{person}{Olivier Chapelle}, \bibinfo{person}{Bernhard
  Scholkopf}, {and} \bibinfo{person}{Alexander Zien}.}
  \bibinfo{year}{2009}\natexlab{}.
\newblock \showarticletitle{Semi-supervised learning (chapelle, o. et al.,
  eds.; 2006)[book reviews]}.
\newblock \bibinfo{journal}{\emph{IEEE Transactions on Neural Networks}}
  \bibinfo{volume}{20}, \bibinfo{number}{3} (\bibinfo{year}{2009}),
  \bibinfo{pages}{542--542}.
\newblock


\bibitem[\protect\citeauthoryear{Chen, Luo, Dong, Li, and He}{Chen
  et~al\mbox{.}}{2018}]%
        {Chen2018FederatedMW}
\bibfield{author}{\bibinfo{person}{Fei Chen}, \bibinfo{person}{Mi Luo},
  \bibinfo{person}{Zhenhua Dong}, \bibinfo{person}{Zhenguo Li}, {and}
  \bibinfo{person}{Xiuqiang He}.} \bibinfo{year}{2018}\natexlab{}.
\newblock \showarticletitle{Federated meta-learning with fast convergence and
  efficient communication}.
\newblock \bibinfo{journal}{\emph{arXiv preprint arXiv:1802.07876}}
  (\bibinfo{year}{2018}).
\newblock


\bibitem[\protect\citeauthoryear{Geyer, Klein, and Nabi}{Geyer
  et~al\mbox{.}}{2017}]%
        {Geyer2017DifferentiallyPF}
\bibfield{author}{\bibinfo{person}{Robin~C Geyer}, \bibinfo{person}{Tassilo
  Klein}, {and} \bibinfo{person}{Moin Nabi}.} \bibinfo{year}{2017}\natexlab{}.
\newblock \showarticletitle{Differentially private federated learning: A client
  level perspective}.
\newblock \bibinfo{journal}{\emph{arXiv preprint arXiv:1712.07557}}
  (\bibinfo{year}{2017}).
\newblock


\bibitem[\protect\citeauthoryear{Grill, Strub, Altch{\'e}, Tallec, Richemond,
  Buchatskaya, Doersch, Pires, Guo, Azar, et~al\mbox{.}}{Grill
  et~al\mbox{.}}{2020}]%
        {grill2020bootstrap}
\bibfield{author}{\bibinfo{person}{Jean-Bastien Grill},
  \bibinfo{person}{Florian Strub}, \bibinfo{person}{Florent Altch{\'e}},
  \bibinfo{person}{Corentin Tallec}, \bibinfo{person}{Pierre~H Richemond},
  \bibinfo{person}{Elena Buchatskaya}, \bibinfo{person}{Carl Doersch},
  \bibinfo{person}{Bernardo~Avila Pires}, \bibinfo{person}{Zhaohan~Daniel Guo},
  \bibinfo{person}{Mohammad~Gheshlaghi Azar}, {et~al\mbox{.}}}
  \bibinfo{year}{2020}\natexlab{}.
\newblock \showarticletitle{Bootstrap your own latent: A new approach to
  self-supervised learning}.
\newblock \bibinfo{journal}{\emph{arXiv preprint arXiv:2006.07733}}
  (\bibinfo{year}{2020}).
\newblock


\bibitem[\protect\citeauthoryear{Han and Zhang}{Han and Zhang}{2020}]%
        {han2020robust}
\bibfield{author}{\bibinfo{person}{Yufei Han} {and} \bibinfo{person}{Xiangliang
  Zhang}.} \bibinfo{year}{2020}\natexlab{}.
\newblock \showarticletitle{Robust Federated Learning via Collaborative Machine
  Teaching.}. In \bibinfo{booktitle}{\emph{AAAI}}. \bibinfo{pages}{4075--4082}.
\newblock


\bibitem[\protect\citeauthoryear{Hard, Rao, Mathews, Ramaswamy, Beaufays,
  Augenstein, Eichner, Kiddon, and Ramage}{Hard et~al\mbox{.}}{2018}]%
        {DBLP:journals/corr/abs-1811-03604}
\bibfield{author}{\bibinfo{person}{Andrew Hard}, \bibinfo{person}{Kanishka
  Rao}, \bibinfo{person}{Rajiv Mathews}, \bibinfo{person}{Swaroop Ramaswamy},
  \bibinfo{person}{Fran{\c{c}}oise Beaufays}, \bibinfo{person}{Sean
  Augenstein}, \bibinfo{person}{Hubert Eichner}, \bibinfo{person}{Chlo{\'e}
  Kiddon}, {and} \bibinfo{person}{Daniel Ramage}.}
  \bibinfo{year}{2018}\natexlab{}.
\newblock \showarticletitle{Federated learning for mobile keyboard prediction}.
\newblock \bibinfo{journal}{\emph{arXiv preprint arXiv:1811.03604}}
  (\bibinfo{year}{2018}).
\newblock


\bibitem[\protect\citeauthoryear{Huang, Yin, Fu, Zhang, Deng, and Liu}{Huang
  et~al\mbox{.}}{2018}]%
        {DBLP:journals/corr/abs-1811-12629}
\bibfield{author}{\bibinfo{person}{Li Huang}, \bibinfo{person}{Yifeng Yin},
  \bibinfo{person}{Zeng Fu}, \bibinfo{person}{Shifa Zhang},
  \bibinfo{person}{Hao Deng}, {and} \bibinfo{person}{Dianbo Liu}.}
  \bibinfo{year}{2018}\natexlab{}.
\newblock \showarticletitle{Loadaboost: Loss-based adaboost federated machine
  learning on medical data}.
\newblock \bibinfo{journal}{\emph{arXiv preprint: 1811.12629}}
  (\bibinfo{year}{2018}).
\newblock


\bibitem[\protect\citeauthoryear{Jeong, Yoon, Yang, and Hwang}{Jeong
  et~al\mbox{.}}{2021}]%
        {Jeong2020FederatedSL}
\bibfield{author}{\bibinfo{person}{Wonyong Jeong}, \bibinfo{person}{Jaehong
  Yoon}, \bibinfo{person}{Eunho Yang}, {and} \bibinfo{person}{Sung~Ju Hwang}.}
  \bibinfo{year}{2021}\natexlab{}.
\newblock \showarticletitle{Federated semi-supervised learning with
  inter-client consistency \& disjoint learning}.
\newblock \bibinfo{journal}{\emph{ICLR}} (\bibinfo{year}{2021}).
\newblock


\bibitem[\protect\citeauthoryear{Jin, Wei, Liu, and Yang}{Jin
  et~al\mbox{.}}{2020}]%
        {jin2020utilizing}
\bibfield{author}{\bibinfo{person}{Yilun Jin}, \bibinfo{person}{Xiguang Wei},
  \bibinfo{person}{Yang Liu}, {and} \bibinfo{person}{Qiang Yang}.}
  \bibinfo{year}{2020}\natexlab{}.
\newblock \bibinfo{title}{Towards Utilizing Unlabeled Data in Federated
  Learning: A Survey and Prospective}.
\newblock
\newblock
\showeprint[arxiv]{2002.11545}~[cs.LG]


\bibitem[\protect\citeauthoryear{Kairouz, McMahan, Avent, Bellet, Bennis,
  Bhagoji, Bonawitz, Charles, Cormode, Cummings, et~al\mbox{.}}{Kairouz
  et~al\mbox{.}}{2019}]%
        {DBLP:journals/corr/abs-1912-04977}
\bibfield{author}{\bibinfo{person}{Peter Kairouz}, \bibinfo{person}{H~Brendan
  McMahan}, \bibinfo{person}{Brendan Avent}, \bibinfo{person}{Aur{\'e}lien
  Bellet}, \bibinfo{person}{Mehdi Bennis}, \bibinfo{person}{Arjun~Nitin
  Bhagoji}, \bibinfo{person}{Keith Bonawitz}, \bibinfo{person}{Zachary
  Charles}, \bibinfo{person}{Graham Cormode}, \bibinfo{person}{Rachel
  Cummings}, {et~al\mbox{.}}} \bibinfo{year}{2019}\natexlab{}.
\newblock \showarticletitle{Advances and open problems in federated learning}.
\newblock \bibinfo{journal}{\emph{arXiv preprint arXiv:1912.04977}}
  (\bibinfo{year}{2019}).
\newblock


\bibitem[\protect\citeauthoryear{Laine and Aila}{Laine and Aila}{2017}]%
        {Laine2017TemporalEF}
\bibfield{author}{\bibinfo{person}{Samuli Laine} {and} \bibinfo{person}{Timo
  Aila}.} \bibinfo{year}{2017}\natexlab{}.
\newblock \showarticletitle{Temporal ensembling for semi-supervised learning}.
\newblock \bibinfo{journal}{\emph{In ICLR, arXiv:1610.02242}}
  (\bibinfo{year}{2017}).
\newblock


\bibitem[\protect\citeauthoryear{LeCun, Bottou, Bengio, and Haffner}{LeCun
  et~al\mbox{.}}{1998}]%
        {lecun1998gradient}
\bibfield{author}{\bibinfo{person}{Yann LeCun}, \bibinfo{person}{L{\'e}on
  Bottou}, \bibinfo{person}{Yoshua Bengio}, {and} \bibinfo{person}{Patrick
  Haffner}.} \bibinfo{year}{1998}\natexlab{}.
\newblock \showarticletitle{Gradient-based learning applied to document
  recognition}.
\newblock \bibinfo{journal}{\emph{Proc. IEEE}} \bibinfo{volume}{86},
  \bibinfo{number}{11} (\bibinfo{year}{1998}), \bibinfo{pages}{2278--2324}.
\newblock


\bibitem[\protect\citeauthoryear{Lee}{Lee}{2013}]%
        {Lee2013PseudoLabelT}
\bibfield{author}{\bibinfo{person}{Dong-Hyun Lee}.}
  \bibinfo{year}{2013}\natexlab{}.
\newblock \showarticletitle{Pseudo-label: The simple and efficient
  semi-supervised learning method for deep neural networks}. In
  \bibinfo{booktitle}{\emph{Workshop on challenges in representation learning,
  ICML}}, Vol.~\bibinfo{volume}{3}.
\newblock


\bibitem[\protect\citeauthoryear{Leroy, Coucke, Lavril, Gisselbrecht, and
  Dureau}{Leroy et~al\mbox{.}}{2019}]%
        {Leroy2019FederatedLF}
\bibfield{author}{\bibinfo{person}{David Leroy}, \bibinfo{person}{Alice
  Coucke}, \bibinfo{person}{Thibaut Lavril}, \bibinfo{person}{Thibault
  Gisselbrecht}, {and} \bibinfo{person}{Joseph Dureau}.}
  \bibinfo{year}{2019}\natexlab{}.
\newblock \showarticletitle{Federated learning for keyword spotting}. In
  \bibinfo{booktitle}{\emph{ICASSP}}. IEEE, \bibinfo{pages}{6341--6345}.
\newblock


\bibitem[\protect\citeauthoryear{Li, Sahu, Zaheer, Sanjabi, Talwalkar, and
  Smith}{Li et~al\mbox{.}}{2018}]%
        {DBLP:journals/corr/abs-1812-06127}
\bibfield{author}{\bibinfo{person}{Tian Li}, \bibinfo{person}{Anit~Kumar Sahu},
  \bibinfo{person}{Manzil Zaheer}, \bibinfo{person}{Maziar Sanjabi},
  \bibinfo{person}{Ameet Talwalkar}, {and} \bibinfo{person}{Virginia Smith}.}
  \bibinfo{year}{2018}\natexlab{}.
\newblock \showarticletitle{Federated optimization in heterogeneous networks}.
\newblock \bibinfo{journal}{\emph{MLSys 2020}} (\bibinfo{year}{2018}).
\newblock


\bibitem[\protect\citeauthoryear{Li, Huang, Yang, Wang, and Zhang}{Li
  et~al\mbox{.}}{2020}]%
        {Li2020OnTC}
\bibfield{author}{\bibinfo{person}{Xiang Li}, \bibinfo{person}{Kaixuan Huang},
  \bibinfo{person}{Wenhao Yang}, \bibinfo{person}{Shusen Wang}, {and}
  \bibinfo{person}{Zhihua Zhang}.} \bibinfo{year}{2020}\natexlab{}.
\newblock \showarticletitle{On the convergence of fedavg on non-iid data}.
\newblock \bibinfo{journal}{\emph{ICLR}} (\bibinfo{year}{2020}).
\newblock


\bibitem[\protect\citeauthoryear{Luping, Wei, and Bo}{Luping
  et~al\mbox{.}}{2019}]%
        {Wang2019CMFLMC}
\bibfield{author}{\bibinfo{person}{WANG Luping}, \bibinfo{person}{WANG Wei},
  {and} \bibinfo{person}{LI Bo}.} \bibinfo{year}{2019}\natexlab{}.
\newblock \showarticletitle{Cmfl: Mitigating communication overhead for
  federated learning}. In \bibinfo{booktitle}{\emph{2019 IEEE 39th
  International Conference on Distributed Computing Systems (ICDCS)}}. IEEE,
  \bibinfo{pages}{954--964}.
\newblock


\bibitem[\protect\citeauthoryear{McMahan, Moore, Ramage, Hampson, and
  y~Arcas}{McMahan et~al\mbox{.}}{2017}]%
        {McMahan2017CommunicationEfficientLO}
\bibfield{author}{\bibinfo{person}{Brendan McMahan}, \bibinfo{person}{Eider
  Moore}, \bibinfo{person}{Daniel Ramage}, \bibinfo{person}{Seth Hampson},
  {and} \bibinfo{person}{Blaise~Aguera y Arcas}.}
  \bibinfo{year}{2017}\natexlab{}.
\newblock \showarticletitle{Communication-efficient learning of deep networks
  from decentralized data}. In \bibinfo{booktitle}{\emph{Artificial
  Intelligence and Statistics}}. PMLR, \bibinfo{pages}{1273--1282}.
\newblock


\bibitem[\protect\citeauthoryear{Miyato, Maeda, Koyama, and Ishii}{Miyato
  et~al\mbox{.}}{2018}]%
        {Miyato2019VirtualAT}
\bibfield{author}{\bibinfo{person}{Takeru Miyato}, \bibinfo{person}{Shin-ichi
  Maeda}, \bibinfo{person}{Masanori Koyama}, {and} \bibinfo{person}{Shin
  Ishii}.} \bibinfo{year}{2018}\natexlab{}.
\newblock \showarticletitle{Virtual adversarial training: a regularization
  method for supervised and semi-supervised learning}.
\newblock \bibinfo{journal}{\emph{IEEE transactions on pattern analysis and
  machine intelligence}} \bibinfo{volume}{41}, \bibinfo{number}{8}
  (\bibinfo{year}{2018}), \bibinfo{pages}{1979--1993}.
\newblock


\bibitem[\protect\citeauthoryear{Park, Park, Shin, and Moon}{Park
  et~al\mbox{.}}{2018}]%
        {park2017adversarial}
\bibfield{author}{\bibinfo{person}{Sungrae Park}, \bibinfo{person}{Jun-Keon
  Park}, \bibinfo{person}{Su-Jin Shin}, {and} \bibinfo{person}{Il-Chul Moon}.}
  \bibinfo{year}{2018}\natexlab{}.
\newblock \showarticletitle{Adversarial dropout for supervised and
  semi-supervised learning}.
\newblock \bibinfo{journal}{\emph{AAAI}} (\bibinfo{year}{2018}).
\newblock


\bibitem[\protect\citeauthoryear{Rasmus, Berglund, Honkala, Valpola, and
  Raiko}{Rasmus et~al\mbox{.}}{2015}]%
        {Rasmus2015SemisupervisedLW}
\bibfield{author}{\bibinfo{person}{Antti Rasmus}, \bibinfo{person}{Mathias
  Berglund}, \bibinfo{person}{Mikko Honkala}, \bibinfo{person}{Harri Valpola},
  {and} \bibinfo{person}{Tapani Raiko}.} \bibinfo{year}{2015}\natexlab{}.
\newblock \showarticletitle{Semi-supervised learning with ladder networks}. In
  \bibinfo{booktitle}{\emph{Advances in neural information processing
  systems}}. \bibinfo{pages}{3546--3554}.
\newblock


\bibitem[\protect\citeauthoryear{Sahu, Li, Sanjabi, Zaheer, Talwalkar, and
  Smith}{Sahu et~al\mbox{.}}{2018}]%
        {Sahu2018OnTC}
\bibfield{author}{\bibinfo{person}{Anit~Kumar Sahu}, \bibinfo{person}{Tian Li},
  \bibinfo{person}{Maziar Sanjabi}, \bibinfo{person}{Manzil Zaheer},
  \bibinfo{person}{Ameet Talwalkar}, {and} \bibinfo{person}{Virginia Smith}.}
  \bibinfo{year}{2018}\natexlab{}.
\newblock \showarticletitle{On the convergence of federated optimization in
  heterogeneous networks}.
\newblock \bibinfo{journal}{\emph{arXiv preprint arXiv:1812.06127}}
  \bibinfo{volume}{3} (\bibinfo{year}{2018}).
\newblock


\bibitem[\protect\citeauthoryear{Sohn, Berthelot, Li, Zhang, Carlini, Cubuk,
  Kurakin, Zhang, and Raffel}{Sohn et~al\mbox{.}}{2020}]%
        {Sohn2020FixMatchSS}
\bibfield{author}{\bibinfo{person}{Kihyuk Sohn}, \bibinfo{person}{David
  Berthelot}, \bibinfo{person}{Chun-Liang Li}, \bibinfo{person}{Zizhao Zhang},
  \bibinfo{person}{Nicholas Carlini}, \bibinfo{person}{Ekin~D Cubuk},
  \bibinfo{person}{Alex Kurakin}, \bibinfo{person}{Han Zhang}, {and}
  \bibinfo{person}{Colin Raffel}.} \bibinfo{year}{2020}\natexlab{}.
\newblock \showarticletitle{Fixmatch: Simplifying semi-supervised learning with
  consistency and confidence}.
\newblock \bibinfo{journal}{\emph{arXiv preprint: 2001.07685}}
  (\bibinfo{year}{2020}).
\newblock


\bibitem[\protect\citeauthoryear{Tarvainen and Valpola}{Tarvainen and
  Valpola}{2017}]%
        {Tarvainen2017MeanTA}
\bibfield{author}{\bibinfo{person}{Antti Tarvainen} {and}
  \bibinfo{person}{Harri Valpola}.} \bibinfo{year}{2017}\natexlab{}.
\newblock \showarticletitle{Mean teachers are better role models:
  Weight-averaged consistency targets improve semi-supervised deep learning
  results}. In \bibinfo{booktitle}{\emph{Advances in neural information
  processing systems}}. \bibinfo{pages}{1195--1204}.
\newblock


\bibitem[\protect\citeauthoryear{Xie, Dai, Hovy, Luong, and Le}{Xie
  et~al\mbox{.}}{2019}]%
        {Xie2019UnsupervisedDA}
\bibfield{author}{\bibinfo{person}{Qizhe Xie}, \bibinfo{person}{Zihang Dai},
  \bibinfo{person}{Eduard Hovy}, \bibinfo{person}{Minh-Thang Luong}, {and}
  \bibinfo{person}{Quoc~V Le}.} \bibinfo{year}{2019}\natexlab{}.
\newblock \showarticletitle{Unsupervised data augmentation for consistency
  training}.
\newblock \bibinfo{journal}{\emph{arXiv preprint arXiv:1904.12848}}
  (\bibinfo{year}{2019}).
\newblock


\bibitem[\protect\citeauthoryear{Yang, Liu, Chen, and Tong}{Yang
  et~al\mbox{.}}{2019a}]%
        {DBLP:journals/corr/abs-1902-04885}
\bibfield{author}{\bibinfo{person}{Qiang Yang}, \bibinfo{person}{Yang Liu},
  \bibinfo{person}{Tianjian Chen}, {and} \bibinfo{person}{Yongxin Tong}.}
  \bibinfo{year}{2019}\natexlab{a}.
\newblock \showarticletitle{Federated machine learning: Concept and
  applications}.
\newblock \bibinfo{journal}{\emph{ACM Transactions on Intelligent Systems and
  Technology (TIST)}} \bibinfo{volume}{10}, \bibinfo{number}{2}
  (\bibinfo{year}{2019}), \bibinfo{pages}{1--19}.
\newblock


\bibitem[\protect\citeauthoryear{Yang, Zhang, Ye, Li, and Xu}{Yang
  et~al\mbox{.}}{2019b}]%
        {Yang2019FFDAF}
\bibfield{author}{\bibinfo{person}{Wensi Yang}, \bibinfo{person}{Yuhang Zhang},
  \bibinfo{person}{Kejiang Ye}, \bibinfo{person}{Li Li}, {and}
  \bibinfo{person}{Cheng-Zhong Xu}.} \bibinfo{year}{2019}\natexlab{b}.
\newblock \showarticletitle{FFD: A Federated Learning Based Method for Credit
  Card Fraud Detection}. In \bibinfo{booktitle}{\emph{International Conference
  on Big Data}}. Springer, \bibinfo{pages}{18--32}.
\newblock


\bibitem[\protect\citeauthoryear{Zhao, Li, Lai, Suda, Civin, and Chandra}{Zhao
  et~al\mbox{.}}{2018}]%
        {DBLP:journals/corr/abs-1806-00582}
\bibfield{author}{\bibinfo{person}{Yue Zhao}, \bibinfo{person}{Meng Li},
  \bibinfo{person}{Liangzhen Lai}, \bibinfo{person}{Naveen Suda},
  \bibinfo{person}{Damon Civin}, {and} \bibinfo{person}{Vikas Chandra}.}
  \bibinfo{year}{2018}\natexlab{}.
\newblock \showarticletitle{Federated learning with non-iid data}.
\newblock \bibinfo{journal}{\emph{arXiv preprint arXiv:1806.00582}}
  (\bibinfo{year}{2018}).
\newblock


\end{thebibliography}

\end{document}